\pdfoutput=1
\RequirePackage{amsthm}
\documentclass[pdflatex,sn-mathphys-num]{sn-jnl}

\usepackage{graphicx}
\usepackage{multirow}
\usepackage{amsmath,amssymb,amsfonts}
\usepackage{amsthm}
\usepackage{mathrsfs}
\usepackage{mathtools}
\usepackage{bm}
\usepackage[title]{appendix}
\usepackage{xcolor}
\usepackage{textcomp}
\usepackage{manyfoot}
\usepackage{booktabs}
\usepackage{placeins}
\usepackage{algorithm}
\usepackage{algorithmicx}
\usepackage{algpseudocode}
\usepackage[shortlabels]{enumitem}

\theoremstyle{thmstyleone}
\newtheorem{theorem}{Theorem}
\newtheorem{proposition}[theorem]{Proposition}
\newtheorem{lemma}[theorem]{Lemma}
\newtheorem{corollary}[theorem]{Corollary}
\theoremstyle{thmstyletwo}

\newtheorem{remark}{Remark}
\theoremstyle{thmstylethree}
\newtheorem{definition}{Definition}

\providecommand{\R}{\mathbb{R}}

\newcommand{\eqdef}{\mathrel{\vcentcolon=}}
\newcommand{\range}{\operatorname{range}}
\newcommand{\tr}{\operatorname{tr}}
\newcommand{\diag}{\operatorname{diag}}
\newcommand{\lse}{\operatorname{lse}}

\newcommand{\rank}{\operatorname{rank}}

\newcommand{\Kexp}{\mathcal{K}_{\exp}}
\newcommand{\Kpsd}{\mathbb{S}_{+}}
\newcommand{\Ksoc}{\mathcal{L}}
\newcommand{\simplex}{\Delta^{n}}
\newcommand{\pstar}{p^{\star}}
\newcommand{\Rpp}{r^{\mathrm{pp}}}
\newcommand{\Rexp}{r^{\exp}}
\newcommand{\cert}{\mathrm{cert}}
\newcommand{\radc}{\operatorname{rad}_c}
\newcommand{\Z}{\mathcal{Z}}
\newcommand{\U}{\mathcal{U}}

\newcommand{\pos}[1]{\left(#1\right)_{+}}

\newcommand{\shi}{s^{\mathrm{hi}}}
\newcommand{\slo}{s^{\mathrm{lo}}}
\newcommand{\ahi}{a^{\mathrm{hi}}}
\newcommand{\alo}{a^{\mathrm{lo}}}
\newcommand{\thi}{t^{\mathrm{hi}}}
\newcommand{\tlo}{t^{\mathrm{lo}}}
\newcommand{\tauhi}{\tau^{\mathrm{hi}}}
\newcommand{\taulo}{\tau^{\mathrm{lo}}}
\newcommand{\hmax}{h^{\max}}
\newcommand{\hmin}{h^{\min}}
\newcommand{\phimin}{\underline{\varphi}}
\newcommand{\KN}{\operatorname{KN}}
\newcommand{\rhorel}{\rho_{\mathrm{rel}}}
\newcommand{\Rlift}{\mathcal{R}^{\exp}_{1}}

\DeclareRobustCommand{\acrown}{\ensuremath{\alpha}\nobreakdash-CROWN}
\makeatletter
\AtBeginDocument{%
\renewenvironment{proof}[1][\proofname]{\par\removelastskip
  \pushQED{\qed}%
  \small\normalfont\topsep7.5\p@\@plus7.5\p@\relax
  \trivlist
  \item[\hskip\labelsep\itshape #1\@addpunct{}]\mbox{}\par\nobreak\noindent\ignorespaces
}{\popQED\endtrivlist\@endpefalse}}
\makeatother

\begin{document}

\title[Conic Certification of Transformer Attention]{Conic Certification of Transformer Attention}

\author*[1]{\fnm{Masoud} \sur{Ataei}}\email{masoud.ataei@utoronto.ca}

\affil*[1]{\orgdiv{Department of Mathematical and Computational Sciences}\\ \orgname{University of Toronto}, \orgaddress{\country{Canada}}}

\abstract{\unboldmath Attention is the computational core of large language models, yet the verifiers that bound how far an attention head's output can move under input perturbations give no guarantee on how loose their bounds are. A weak certificate therefore cannot be told apart from a fragile model. We develop a conic programming theory of attention certification that supplies this guarantee. We prove that the difficulty lies not in the softmax but in the bilinear interaction of a varying query with the keys. This interaction makes certification NP-hard over boxes and inapproximable once the attention logits can move by a constant amount. A fixed query or listed token substitutions, by contrast, make the exact worst case efficiently computable. An exponential cone moment relaxation then certifies a head with an error bounded quadratically in the perturbation radius by computable quantities, an order that cannot be improved. We also show that relaxations that treat a network one operation at a time, linear bound propagation among them, pay on second-order terms a sum of per-operation maxima instead of one coupled maximum. Computational results on GPT-2 indicate that the theory carries over to a trained model. Against exact worst cases with the query fixed, the certificate attains the predicted second-order accuracy. It is tighter than linear bound propagation on every head, and the looseness of bound propagation tracks the predicted per-operation excess.}


\maketitle

\clearpage
\section{Introduction}\label{sec:intro}

Transformers~\cite{vaswani2017attention} have become the computational substrate of modern
artificial intelligence, and how their behaviour can change under a change of their inputs
has become a central question of trustworthy artificial intelligence. Adversarial
examples~\cite{szegedy2014intriguing,goodfellow2015explaining}, and their counterparts for
language models~\cite{zou2023universal}, show that small and deliberately chosen changes can
redirect what a model computes. A certificate answers the question for all inputs at once.
Given a set of admissible inputs and a quantity read linearly from the model, such as a logit
margin, it returns a number that the quantity provably cannot exceed anywhere in the set. A
certificate is sound when it never underestimates the worst case, and it is useful only to the
extent that it does not overestimate it by much.

For networks of piecewise-linear units, certification is mature. Complete methods solve the
problem exactly by satisfiability modulo theories, mixed-integer programming or
branch-and-bound~\cite{katz2017reluplex,tjeng2019evaluating,wang2021beta}. Incomplete methods
relax it by linear programming, bound propagation, abstract interpretation or semidefinite
programming~\cite{wong2018provable,zhang2018efficient,gowal2019scalable,singh2019abstract,raghunathan2018semidefinite,xu2020automatic},
and the limits of relaxing one neuron at a time are relatively well
understood~\cite{salman2019convex}. Attention is different. It couples a bilinear interaction
between tokens with the softmax normalization. The verifiers built for it propagate linear bounds
or zonotopes through the two operations separately~\cite{shi2020robustness,bonaert2021fast,huang2026parameterized},
with convex bounds on the softmax itself~\cite{wei2023convex}, or work at the level of the
text~\cite{jia2019certified,kumar2024certifying}, and randomized
smoothing~\cite{cohen2019certified} offers probabilistic guarantees of another kind. For sparsemax
attention, mixed-integer programming finds the exact worst case~\cite{liao2023transformers}, and
Rezazadeh and Gholami Davoodi~\cite{rezazadeh2026vertex} compute the tightest bound on the softmax
that score intervals alone allow. These verifiers are sound and can be fast. None of them, however, comes with a guarantee on how far its
bound lies from the true worst case or on how that distance scales with the perturbation, and the
complexity of the underlying problem has not been mapped. A loose certificate therefore cannot be
attributed either to the method or to the model.

This paper develops a theory of attention certification based on conic programming. Its object
is the central primitive of a transformer, namely one attention head with fixed weights, input
tokens varying in convex sets, and a linear functional of the head's output. The construction
rests on one observation. The softmax, usually regarded as the obstacle, has a rigid structure.
The exponential cone represents the exponential in it exactly from below, and a secant bounds it from above with a slack known in closed form. The difficulty lies instead in the bilinear score between a query and
a key. Lifting the problem into a moment matrix meets this difficulty by coupling where a head
attends with what it reads.

The ingredients are classical, and the novelty lies in what is proved about them. They are
moment relaxations of nonconvex quadratic
problems~\cite{shor1987quadratic,lasserre2001global,sherali1990hierarchy}, the exponential cone
of relative entropy programming~\cite{chandrasekaran2017relative}, the hardness of the cut
norm~\cite{alon2006approximating} and Dinkelbach's method of fractional
programming~\cite{dinkelbach1967nonlinear}. Semidefinite certificates for ReLU
networks~\cite{raghunathan2018semidefinite}, curvature-based certificates for smooth
networks~\cite{singla2020second} and Lipschitz analyses of attention~\cite{kim2021lipschitz}
neither relax the softmax nor establish an order of accuracy in the perturbation radius. What
this paper contributes is an accuracy theory for the coupled problem. It proves that the moment
relaxation is accurate to second order in the perturbation radius, with constants that do not
depend explicitly on the sequence length or the dimension, and that this order is attained, so it
cannot be improved. It also identifies the computable quantity at the core of the error. Semidefinite relaxations of
quadratic problems that are exact at a nominal instance can remain exact under small changes of the
data~\cite{cifuentes2022local}. Here, the admissible set grows from a point, and the gap of the
relaxation is of exact second order. The paper also maps the complexity of the problem, with exact algorithms against which certificates are measured. Moreover, in the spirit of the convex barrier of Salman et
al.~\cite{salman2019convex} but at second order, it identifies the loss that relaxations treating
one operation at a time cannot avoid.

The results form three strata. The first locates the difficulty. The softmax does not make
certification hard. Its Lipschitz and interval bounds do not grow with the context or the
dimension, and relaxed attention patterns obey an exact identity that weights their slack by the
values the head reads. Hardness comes from the bilinear score alone. With a varying query and
tokens in boxes, certification is NP-hard already for two tokens, and inapproximable once the
attention logits can move by a constant amount. With a fixed query, the exact worst case is the
root of a separable scalar equation, computed in polynomial time for boxes and ellipsoids, and
listed token substitutions are solved exactly for any query.

The second stratum constructs the certificates. A linear program over the admissible attention
patterns is solved greedily, with a first-order gap that does not depend explicitly on the dimension. Its solution exhibits a worst admissible attention pattern, and its sensitivities show how the bound responds to the values of the keys and to the attention bounds. A level-one
exponential cone moment relaxation has a gap bounded by a value-weighted secant slack. This slack
vanishes quadratically in the score width and hence in the perturbation radius. The gap of the
relaxation is of exact second order already for a head with two keys and a fixed query, in which one
key varies in a disc. The radius certified to a given accuracy therefore grows like the
square root of the tolerance. Safe dual bounds with an explicit rounding allowance, and an exact reduction to each head's
subspace, make the certificate robust to solver inaccuracy and affordable on a trained model.

The third stratum explains the looseness of bound propagation. Relaxations that bound one
operation at a time include linear bound propagation with any choice of slopes. On second-order
parts and remainders, they pay a sum of per-operation second-order maxima, whereas the exact term
is one coupled maximum. The excess grows with the effective rank of the coupled term.

The theory is tested on synthetic heads and on all heads of GPT-2
small~\cite{radford2019language}, with perturbations of the normalized tokens that each head
reads. Against exact worst cases with the query fixed, the gap of the second-order
certificate grows quadratically in the perturbation radius, as predicted. The certificate is
tighter than the linear bound propagation methods CROWN~\cite{zhang2018efficient} and
\acrown{}~\cite{xu2021fast} on every head, typically by a factor of four to twenty. On the
remainders of GPT-2 blocks, the looseness of CROWN tracks the predicted per-operation excess and
exceeds it only by a small factor. The code and the exact worst cases are publicly
available.\footnote{\url{https://github.com/chimera1001/conic-attention-cert}}\enlargethispage{3pt}

The paper is organized as follows. Section~\ref{sec:problem} sets up the problem, the quantities
that control it and the structure of the softmax, and Section~\ref{sec:complexity} maps the
complexity of certification. Section~\ref{sec:firstorder} constructs the two certificates and
Section~\ref{sec:computation} describes how they are computed. Section~\ref{sec:perop} analyses
per-operation relaxations, Section~\ref{sec:experiments} reports the computational results and
Section~\ref{sec:conclusion} concludes. Proofs are collected in Appendix~\ref{app:proofs}, and the
relaxation is written out in Appendix~\ref{app:relaxation}.

\section{Foundations}\label{sec:problem}

A certificate is a statement about every admissible input at once, and what it can
promise depends on how precisely the problem it bounds is posed. This section fixes the object of the paper, one attention head read along one direction. It recalls the conic programs used to certify it and introduces the quantities, computed before any optimization, that control the accuracy of every certificate. It then establishes the structure of the softmax on which the certificates rest. Throughout, $\|\cdot\|$ is the
Euclidean norm of a vector and the spectral norm of a matrix,
$\range(v)=\max_kv_k-\min_kv_k$, $\simplex$ is the probability simplex in $\R^{n}$,
$\sigma$ is the softmax, $\lse(s)=\log\sum_le^{s_l}$ and $\pos{x}=\max\{x,0\}$.

A transformer processes a sequence of tokens $x_1,\dots,x_n\in\R^{d}$, the
\emph{residual stream}. In the pre-normalization architecture of GPT-2, each block adds to the stream the outputs of $H$ attention heads and then of a multilayer perceptron, so that
\begin{equation}\label{eq:block}
  x'_j=x_j+\sum_{h=1}^{H}y^{(h)}_j\bigl(\mathrm{LN}_1(x_1),\dots,\mathrm{LN}_1(x_n)\bigr),
  \qquad
  x''_j=x'_j+\mathrm{MLP}\bigl(\mathrm{LN}_2(x'_j)\bigr),
\end{equation}
where $\mathrm{LN}$ denotes layer normalization~\cite{ba2016layer}, defined by
\begin{equation}\label{eq:layernorm}
  \mathrm{LN}(x)=g\odot\frac{Px}{\bigl(\|Px\|^{2}/d+\epsilon_{\mathrm{LN}}\bigr)^{1/2}}+b,
\end{equation}
where $P=I-\tfrac1d\mathbf 1\mathbf 1^\top$ is the centring projection and $g$ and $b$ are a learned gain and bias. The perceptron applies the Gaussian error linear unit~\cite{hendrycks2016gaussian} between two affine maps. A head maps the normalized tokens $z_j=\mathrm{LN}_1(x_j)$ to queries, keys
and values through matrices $W_Q,W_K\in\R^{d\times d_k}$ and $W_V\in\R^{d\times d_v}$ and
biases $b_Q,b_K,b_V$, and writes back through $W_O\in\R^{d_v\times d}$. At a query
position $i$ with a causal mask, it attends to the positions $j\le i$ with scores and weights
\begin{equation}\label{eq:scores}
  s_j=\frac{(W_Q^\top z_i+b_Q)^\top(W_K^\top z_j+b_K)}{\sqrt{d_k}},
  \qquad
  a=\sigma(s),
  \qquad
  \sigma_j(s)=\frac{e^{s_j}}{\sum_{l}e^{s_l}},
\end{equation}
and its output is
\begin{equation}\label{eq:headout}
  y_i=W_O^\top\sum_{j\le i}a_j\bigl(W_V^\top z_j+b_V\bigr).
\end{equation}
In the language of mechanistic interpretability~\cite{elhage2021mathematical}, the head
composes a query--key circuit, which decides where to attend, with an output--value
circuit, which decides what is written. Without biases the two circuits are, respectively, a bilinear and a linear map of the tokens, namely
\begin{equation}\label{eq:MN}
  s_j=z_i^\top Mz_j,\quad M=\frac{W_QW_K^\top}{\sqrt{d_k}},
  \qquad
  y_i=\sum_ja_jNz_j,\quad N=(W_VW_O)^\top.
\end{equation}
Biases are absorbed by appending a unit coordinate to every token, so every statement below is made for the bias-free form~\eqref{eq:MN} and holds verbatim for affine scores and values, with all norms and constants computed in the extended coordinates. GPT-2
small has $12$ blocks of $H=12$ heads, $d=768$ and
$d_k=d_v=64$~\cite{radford2019language}.

The certification problem concerns one head with fixed matrices $M$ and $N$, a query
position $i$ and the keys $j\in J$, $|J|=n$, that it attends to; in a causal transformer
the query is itself one of the keys. Each token varies in a compact convex set, written in \emph{base coordinates} as
\begin{equation}\label{eq:sets}
  \Z_j=\bigl\{\bar z_j+B_ju:\ u\in\U_j\bigr\},
\end{equation}
where $\bar z_j$ is the nominal token, $B_j\in\R^{d\times r_j}$ and $\U_j$ is a ball $\{\|u\|\le\rho_j\}$, a box $\{|u|\le\omega_j\}$ or the point $\{0\}$ of a fixed token. A ball makes $\Z_j$ an ellipsoid, and the radius of a box is its circumradius $\rho_j=\|\omega_j\|$.
For an input $z=(z_j)_{j\in J}$, the scores, the pattern and the output are, respectively,
\begin{equation}\label{eq:head}
  s_j(z)=z_i^\top Mz_j,
  \qquad
  a(z)=\sigma\bigl(s(z)\bigr)\in\simplex,
  \qquad
  y(z)=\sum_{j\in J}a_j(z)\,Nz_j .
\end{equation}
A \emph{specification} is a vector $c\in\R^{d}$, and the quantity to be certified is $\varphi(z)=c^\top y(z)$. It is the projection of the head's output on a probe or steering direction or, with $c$ the difference of two rows of the unembedding matrix, the direct contribution of the head to a logit margin, with the scale of the final layer normalization frozen.

\begin{definition}[Head certification problem]\label{def:problem}
The worst case of the specification over the admissible inputs is
\begin{equation}\label{eq:pstar}
  \pstar=\max\bigl\{\varphi(z):\ z_j\in\Z_j\ \text{for all }j\in J\bigr\}.
\end{equation}
A \emph{certificate} is a number $r\ge\pstar$ computed by a procedure that is valid for
every instance, and its \emph{gap} is $r-\pstar$.
\end{definition}

Lower bounds come from the same problem with $-c$. Problem~\eqref{eq:pstar} is nonconvex for two reasons that the rest of the paper separates. The scores are bilinear in the
tokens whenever the query varies, and the pattern is a nonlinear function of the scores
that multiplies the values. Because the residual connection adds the output of each head
linearly, the contribution of one head to the residual stream, or to a quantity read linearly from it, is an instance of the head certification problem~\eqref{eq:pstar}.

Accuracy is measured by how the gap behaves as the perturbation shrinks. When the sets
are scaled by a radius $\rho$, a certificate is of \emph{first order} if its gap is $O(\rho)$ and of
\emph{second order} if it is $O(\rho^{2})$. The output moves by $\Theta(\rho)$, so a
first-order certificate overestimates the worst case by a fixed fraction of the output's
range of motion however small the perturbation. A second-order certificate, by contrast, becomes exact relative to that range as $\rho\to0$. Equivalently, the largest radius at which a
certificate is within a tolerance $\varepsilon$ of the truth grows like $\varepsilon$ for
the first and like $\sqrt{\varepsilon}$ for the second. Interval bounds are of first order
for attention, and for the linear bound propagation and zonotope verifiers of
transformers no order of accuracy has been established.

The certificates are values of conic programs. Given a proper cone $\mathcal K\subset\R^{m}$ and data $A$, $b$ and $f$ of compatible sizes, the primal and dual programs are, respectively,
\begin{equation}\label{eq:conic}
    \max_{x}\ \bigl\{f^\top x:\ b-Ax\in\mathcal K\bigr\},
  \qquad
  \min_{\lambda}\ \bigl\{b^\top\lambda:\ A^\top\lambda=f,\ \lambda\in\mathcal K^{*}\bigr\},
\end{equation}
with dual cone $\mathcal K^{*}=\{\lambda:\lambda^\top w\ge0\ \forall w\in\mathcal K\}$. They satisfy weak duality without assumptions, so any dual feasible point certifies an upper bound~\cite{bental2001lectures}. Four cones are used, namely the nonnegative orthant, the
second-order cone $\Ksoc=\{(t,w):\|w\|\le t\}$, the cone $\Kpsd$ of positive semidefinite
matrices and the exponential cone
\begin{equation}\label{eq:kexp}
  \Kexp=\operatorname{cl}\bigl\{(x_1,x_2,x_3)\in\R^{3}:\ x_2>0,\ x_1\ge x_2e^{x_3/x_2}\bigr\},
\end{equation}
which interior-point methods handle in polynomial
time~\cite{dahl2022primal,chandrasekaran2017relative}. The constraint
$(a,1,t)\in\Kexp$ states $a\ge e^{t}$. This is why the exponential inside the softmax can be represented exactly in a conic relaxation. Nonconvex quadratic problems in a vector $\xi$ are relaxed by \emph{lifting}, in which a symmetric matrix $Y$ replaces the products
$\xi\xi^\top$ in the moment matrix
\begin{equation}\label{eq:moment}
  X=\begin{pmatrix}1&\bar\xi^\top\\ \bar\xi& Y\end{pmatrix}\in\Kpsd ,
\end{equation}
so that every quadratic expression becomes linear in $X$. This is the Shor
relaxation~\cite{shor1987quadratic}, the first level of the moment hierarchy~\cite{lasserre2001global}. Products of valid constraints expressed in $X$ tighten it, as in the reformulation--linearization technique~\cite{sherali1990hierarchy}.
For a quadratic objective over one ball the Shor relaxation is exact by the
S-lemma~\cite{polik2007survey}, and for bilinear objectives over boxes it is exact up to the
Grothendieck constant~\cite{alon2006approximating}.

The accuracy of every certificate below is governed by a few quantities computed from the
instance before any optimization, collected in Table~\ref{tab:controls}. The first step
bounds each score over the whole admissible set by certified intervals
\begin{equation}\label{eq:scoreint}
  \slo_j\le s_j(z)\le\shi_j\qquad\text{for all }z\in\Z_1\times\cdots\times\Z_n .
\end{equation}
Their widths $\Delta_j=\shi_j-\slo_j$ and the \emph{score width} $\Delta=\max_j\Delta_j$
measure how far the attention logits can swing. The midpoints
$s^{\mathrm{mid}}_j=(\slo_j+\shi_j)/2$ define the \emph{midpoint pattern} $\hat a=\sigma(s^{\mathrm{mid}})$. Its \emph{dispersion} and the \emph{margin} are, respectively,
\begin{equation}\label{eq:gamma}
  \theta=1-\|\hat a\|^{2},
  \qquad
  \gamma=\max_{j^\star}\Bigl(\slo_{j^\star}-\max_{j\ne j^\star}\shi_j\Bigr).
\end{equation}
The dispersion lies in $(0,1-1/n]$ for $n\ge2$, small when $\hat a$ is close to one-hot and largest when it is uniform. It enters none of the bounds and serves to describe how concentrated the attention of a head is. The margin, when positive, certifies that one key dominates all others on the whole set.

From the score intervals follow certified intervals $\alo_j\le a_j\le\ahi_j$ for the attention weights in closed form. They also give an interval $[\taulo,\tauhi]$ for the normalizer $\tau=\lse(s)$ and intervals for the \emph{log-attention} $t_j=s_j-\tau=\log a_j$, namely
\begin{equation}\label{eq:tint}
  \taulo=\lse(\slo),
  \qquad
  \tauhi=\lse(\shi),
  \qquad
  \tlo_j=\log\alo_j,
  \qquad
  \thi_j=\log\ahi_j,
\end{equation}
of widths $w_j=\thi_j-\tlo_j$ and maximal width $w_t=\max_jw_j$, the \emph{secant width}.
Since $\log\alo_j\ge\slo_j-\tauhi$ and $\log\ahi_j\le\shi_j-\taulo$, the widths satisfy
\begin{equation}\label{eq:wbound}
  w_j\le\Delta_j+\tauhi-\taulo\le2\Delta .
\end{equation} The second-order certificate represents
$a_j\ge e^{t_j}$ exactly and $a_j\le e^{t_j}$ by the secant $\ell_j$ of the exponential on
$[\tlo_j,\thi_j]$. The slack of this representation is measured by the \emph{exact secant
gap}, the \emph{weighted secant slack} and the \emph{slack budget}, defined respectively by
\begin{equation}\label{eq:secant}
  D_j=\max_{t\in[\tlo_j,\thi_j]}\bigl(\ell_j(t)-e^{t}\bigr),
  \qquad
  S=\sum_jS_j\ \ \text{with}\ \ S_j=\frac{\ahi_j\,w_j^{2}}{8},
  \qquad
  D_{\mathrm{tot}}=1-\sum_j\alo_j .
\end{equation}

The values enter through their projection on the specification. Let $h_j(z)=c^\top Nz$ be
the value of token $j$ along $c$ and $g_j=B_j^\top N^\top c$ its gradient in base coordinates. Let $\hmax_j$ and $\hmin_j$ be, respectively, the maximum and the minimum of $h_j$ over $\Z_j$,
let $\nu_{c,j}=\max_{u\in\U_j}|g_j^\top u|$, the \emph{directional value width}, which equals $\rho_j\|g_j\|$ for a ball, and let $\nu_c=\max_j\nu_{c,j}$. The \emph{directional value radius}
\begin{equation}\label{eq:radc}
  \radc=\tfrac12\Bigl(\max_j\hmax_j-\min_j\hmin_j\Bigr)
\end{equation}
is the half-range of the values along $c$. It is at most $\|c\|$ times the Chebyshev radius
of the union of the value sets $N\Z_j$, and much smaller when the values spread in
directions orthogonal to $c$, as they do in trained heads. The bilinear part
$u_i^\top B_i^\top MB_ju_j$ of the score of key $j$ is bounded by the \emph{score
covariance}
\begin{equation}\label{eq:eta}
  \eta=\max_{j\in J}\ \|B_i^\top MB_j\|\,\rho_i\rho_j ,
\end{equation}
which vanishes for a fixed query and is of order $\rho^{2}$ in general. The last quantity
combines the slack of the softmax with the values. Let $\phimin$ be any certified lower
bound on $\min_z\varphi(z)$, for instance $\min_j\hmin_j$, since the output is a convex combination
of the values. The \emph{value-weighted secant slack} is the value of a fractional knapsack, namely
\begin{equation}\label{eq:KN}
\begin{aligned}
  \KN(\chi,D,D_{\mathrm{tot}})&=\max\Bigl\{\sum_jd_j\chi_j:\ 0\le d_j\le D_j,\ \sum_jd_j\le D_{\mathrm{tot}}\Bigr\},\\
  \chi_j&=\pos{\hmax_j-\phimin+\eta\,\radc},
\end{aligned}
\end{equation}
obtained by filling the keys in decreasing order of $\chi_j$. It measures how much
attention mass a relaxation can misplace, weighted by how far the value of each key can exceed the output. It is the central quantity of the paper.

\begin{table}[t]
\caption{Controlling quantities of a head certification instance, all computed in closed
form or by small convex programs before the certificate is solved, with their orders as
$\rho\to0$ for balls of radius $\rho$.}
\label{tab:controls}
\begin{tabular*}{\textwidth}{@{\extracolsep\fill}lll}
\toprule
symbol & meaning & order\\
\midrule
$\Delta$ & score width, $\max_j(\shi_j-\slo_j)$ & $\rho$\\
$w_t$ & secant width, $\max_j\log(\ahi_j/\alo_j)\le2\Delta$ & $\rho$\\
$\theta,\ \gamma$ & dispersion of the midpoint pattern, margin of the dominant key & $1$\\
$\alo_j,\ahi_j$ & attention intervals & $\ahi_j-\alo_j=O(\rho)$\\
$S,\ D_j$ & weighted secant slack, exact secant gap~\eqref{eq:secant} & $\rho^{2}$\\
$\radc,\ \nu_c$ & directional value radius and width & $1$, $\rho$\\
$\eta$ & score covariance, the bilinear part of the scores~\eqref{eq:eta} & $\rho^{2}$\\
$\KN$ & value-weighted secant slack~\eqref{eq:KN} & $\rho^{2}$\\
\botrule
\end{tabular*}
\end{table}

Gaps are reported on two scales. The \emph{relative gap} divides the gap by the output's
range of motion $\max\varphi-\min\varphi$ over the admissible set, in practice by a certified or attained lower bound on it. It compares the certificate with what the perturbation can do. The \emph{range-normalized gap} divides by the nominal spread $\max_jh_j(\bar z_j)-\min_jh_j(\bar z_j)$ of the values along $c$ and compares the
certificate with the scale of the head's output.

The softmax is often regarded as what makes attention hard to certify. It has, on the contrary, a rigid structure. Its constants do not depend on the number of keys or the
dimension, its interval bounds are explicit, and the patterns produced by an
exponential cone relaxation obey an exact identity. Adding a constant to all scores leaves
the softmax unchanged, so the natural measure of a change of scores is its range.

\begin{lemma}[Lipschitz bound]\label{lem:lip}
For all $s,s'\in\R^{n}$,
\begin{equation}\label{eq:lip}
  \|\sigma(s)-\sigma(s')\|_1\le\tfrac12\range(s-s')\le\|s-s'\|_\infty ,
\end{equation}
and the constant $\tfrac12$ cannot be lowered.
\end{lemma}

The proof (Appendix~\ref{app:proofs}) rests on one fact. The Jacobian of the softmax maps a change $u$ of the scores to a vector whose $\ell_1$ norm is the mean absolute deviation of $u$ under the pattern, and this deviation is at most half the range of $u$. The total variation distance between two patterns is thus at most a quarter of the largest
change in the difference of two logits. The bound implies the $\ell_1$ case of the bound of
Nair~\cite{nair2026softmax}, since $\range(v)\le\|v\|_1$, and its range form is the one that
certified score intervals control directly. These intervals determine the attention
intervals.

\begin{lemma}[Attention intervals]\label{lem:intervals}
Let the scores lie in $\mathcal B=\prod_j[\slo_j,\shi_j]$, with maximal width $\Delta$ and
midpoint pattern $\hat a$. Then, the following hold.
\begin{enumerate}[label=\normalfont(\roman*), leftmargin=*]
\item For every $j$, $\sigma_j(\mathcal B)=[\alo_j,\ahi_j]$ with
\begin{equation}\label{eq:aint}
  \alo_j=\Bigl(1+\sum_{l\ne j}e^{\shi_l-\slo_j}\Bigr)^{-1},
  \qquad
  \ahi_j=\Bigl(1+\sum_{l\ne j}e^{\slo_l-\shi_j}\Bigr)^{-1}.
\end{equation}
\item For every $j$, and for the sums over the keys,
\begin{equation}\label{eq:Wbound}
\begin{gathered}
  e^{-\Delta}\hat a_j\le\alo_j\le\ahi_j\le e^{\Delta}\hat a_j,\\
  \sum_j\ahi_j\le e^{\Delta},\\
  W\eqdef\sum_j\bigl(\ahi_j-\alo_j\bigr)\le2\sinh\Delta .
\end{gathered}
\end{equation}
\item If a key $j^\star$ has margin $\gamma>0$, then for every $s\in\mathcal B$,
\begin{equation}\label{eq:margin}
  \|\sigma(s)-e_{j^\star}\|_1\le\frac{2(n-1)e^{-\gamma}}{1+(n-1)e^{-\gamma}} .
\end{equation}
\end{enumerate}
\end{lemma}

No key can gain or lose more than a factor $e^{\Delta}$ of its midpoint attention, so the total uncertainty of the pattern is of order $\Delta$. With a margin, the pattern is within
an amount exponentially small in $\gamma$ of one-hot. The slack of the secant
representation of the exponential is known exactly.

\begin{lemma}[Exact secant gap]\label{lem:secant}
Let $\ell$ be the secant of the exponential on an interval $[\tlo,\thi]$ of width $w>0$, and
put
\begin{equation}\label{eq:Dclosed}
  y=\frac{e^{w}-1}{w}-1,
  \qquad
  D=e^{\tlo}\bigl((1+y)\log(1+y)-y\bigr).
\end{equation}
Then, $\max_{[\tlo,\thi]}(\ell(t)-e^{t})=D$, with
\begin{equation}\label{eq:Dbounds}
  D\le\min\Bigl\{\tfrac{w^{2}}{8}e^{\thi},\ e^{\thi}-e^{\tlo}\Bigr\},
\end{equation}
and, as $w\to0$,
\begin{equation}\label{eq:Dasym}
  D=\tfrac{w^{2}}{8}e^{\tlo}\bigl(1+O(w)\bigr).
\end{equation}
\end{lemma}

The classical interpolation bound $\frac{w^{2}}{8}e^{\thi}$ is accurate for narrow intervals,
but for wide ones the exact gap saturates at $e^{\thi}-e^{\tlo}$. Since $e^{\thi_j}=\ahi_j$,
the lemma gives $D_j\le\ahi_jw_j^{2}/8$, so $\sum_jD_j\le S$. The weighted slack $S$ is the transparent quantity, of order $\Delta^{2}$, and $D_j$ the sharp one.

The structural fact on which the second-order certificate rests concerns relaxed patterns. A
convex relaxation cannot enforce $a_j=e^{t_j}$; it enforces only the sandwich
$e^{t_j}\le a_j\le\ell_j(t_j)$ together with $\sum_ja_j=1$. How such a relaxed pattern can
differ from a genuine softmax pattern in its effect on the objective is described by an exact
identity.

\begin{proposition}[Value-weighted sandwich identity]\label{prop:P7v}
Let $a_j\ge e^{t_j}$ for all $j$, $\sum_ja_j=1$, $\sigma=\sigma(t)$ and $d_j=a_j-e^{t_j}$.
Then, for every $h\in\R^{n}$,
\begin{equation}\label{eq:P7v}
  \sum_j\bigl(a_j-\sigma_j\bigr)h_j=\sum_jd_j\bigl(h_j-\bar h_\sigma\bigr),
\end{equation}
where $\bar h_\sigma=\sum_l\sigma_lh_l$ is the average of $h$ under $\sigma$, and $\sum_jd_j=1-\sum_je^{t_j}$.
If moreover $t_j\in[\tlo_j,\thi_j]$ and $a_j\le\ell_j(t_j)$ for all $j$, then
\begin{equation}\label{eq:P7vKN}
  \sum_j\bigl(a_j-\sigma_j\bigr)h_j\le\KN\bigl(\pos{h-\bar h_\sigma},D,D_{\mathrm{tot}}\bigr).
\end{equation}
\end{proposition}

The relaxation can inflate the attention of key $j$ above the value that the exponential
dictates by at most $D_j$, and the normalization removes the inflated mass from all keys in
proportion to the genuine pattern $\sigma$. The net effect on the objective is the inflation of each key times the excess of its value over the average $\bar h_\sigma$. Only keys whose values lie above that average can therefore increase the objective, however much slack they carry. The knapsack bound~\eqref{eq:P7vKN} needs no smallness condition and saturates automatically, since
$D_j\le e^{\thi_j}-e^{\tlo_j}$ and $D_{\mathrm{tot}}\le1$. A key whose value lies below
$\bar h_\sigma$ contributes nothing to it. This is why a measure of slack that ignores the values, such as $S$ alone, cannot predict the accuracy of the certificate on trained heads,
where most of the slack sits on keys with small attention and low values.

\section{Complexity}\label{sec:complexity}

The softmax by itself is not a source of difficulty. Its Lipschitz and interval bounds do not depend on the number of keys, and with scores in a box its worst case is found by sorting. This section locates the difficulty of computing the worst case $\pstar$ of the head certification problem~\eqref{eq:pstar}. The problem is NP-hard as soon as the query and a key vary jointly in boxes, already for two tokens. It cannot be approximated within a constant fraction of the output range when the attention logits can move by a constant amount. With a fixed query it reduces to a scalar root-finding problem, solved exactly in polynomial time for boxes, zonotopes, convex hulls of listed points and ellipsoids. Listed token substitutions are solved exactly for any query. Together, these
results determine which guarantees a polynomial-time certificate can and cannot offer.

The bilinear score alone carries the hardness. The construction uses fixed values, no
multilayer perceptron and only two tokens, one of which is the query.

\begin{theorem}[NP-hardness with two tokens]\label{thm:hard}
For one head with two tokens varying in boxes, one of them the query, and values that do
not depend on the tokens, deciding whether $\pstar\ge\beta$ is NP-hard.
\end{theorem}

The reduction encodes the infinity-to-one norm of a matrix $Q$, namely
\begin{equation*}
  \|Q\|_{\infty\to1}=\max\bigl\{u^\top Qu':\ u\in\{\pm1\}^{d_1},\ u'\in\{\pm1\}^{d_2}\bigr\},
\end{equation*}
into the score between the two tokens, so that the output is a strictly increasing function of $u^\top Qu'$. For a matrix built from the edges of a graph, this norm equals four times the maximum cut of the graph~\cite{alon2006approximating}, so deciding a threshold for it is NP-hard. The full construction is given in Appendix~\ref{app:proofs}. When the scores themselves, instead of the tokens, range over a box, the
problem is easy.

\begin{remark}[Scores in a box]\label{rem:threshold}
For fixed values $v$ and scores in a box, the maximum of $v^\top\sigma(s)$ is attained at
one of the $n+1$ vertices obtained by sorting $v$ and raising the scores of a top segment,
as also observed by Rezazadeh and Gholami Davoodi~\cite{rezazadeh2026vertex}. At a maximizer, every key whose value exceeds the maximum has its score at the upper end, and every key whose value falls below it has its score at the lower end. This holds because the derivative of the objective in the score of key $j$ is $\sigma_j(s)$ times the difference between $v_j$ and the objective, and the scores of keys whose value equals the maximum do not affect it.
The maximum is computed in $O(n\log n)$ time.
\end{remark}

The hardness is therefore created by the map from tokens to scores, not by the softmax.
Since the infinity-to-one norm is also hard to approximate, the reduction transfers
inapproximability. The error is measured relative to the range $\max\varphi-\min\varphi$ of the output over the admissible set. In the reduction, the score between the two tokens is divided by an estimate of $\|Q\|_{\infty\to1}$ that the semidefinite relaxation of Grothendieck's inequality computes in polynomial time to within a factor $1.785$ (Appendix~\ref{app:proofs}). The score then ranges over an interval of length between $2/1.785>1.12$ and $2$, so the attention logits move by a constant amount, neither small nor large.

\begin{theorem}[Inapproximability at constant score width]\label{thm:inapprox}
Consider heads with two tokens varying in boxes, one of them the query, and values that do not
depend on the tokens. Unless $\mathrm{P}=\mathrm{NP}$, no polynomial-time algorithm returns, for every
such head whose score between the two tokens ranges over an interval of length in $[1.12,2]$, a
number $r$ with
\begin{equation}\label{eq:inapprox}
  \pstar\le r\le\pstar+\tfrac{1}{200}\bigl(\max\varphi-\min\varphi\bigr).
\end{equation}
\end{theorem}

The theorem concerns every polynomial-time method, not a particular relaxation.
Second-order guarantees become vacuous when the score width is of order one, and over boxes no
efficient method can extend them to that regime. When the query varies and the logits can swing by a constant while several keys compete, certificates are necessarily loose on some instances.
The constant $1/200$, which is not optimized, is small enough that an algorithm meeting it would approximate the maximum cut beyond H\aa stad's threshold of $16/17$~\cite{hastad2001some}, as the proof in Appendix~\ref{app:proofs} shows.

\begin{remark}\label{rem:ugc}
Under the Unique Games Conjecture, $\|Q\|_{\infty\to1}$ cannot be approximated in polynomial
time within any factor below the Grothendieck constant
$K_G\in(1.676,1.783)$~\cite{raghavendra2009towards,davie1984lower,reeds1991new,krivine1979constantes}. Even the largest score that two tokens can produce therefore cannot be estimated better than by the
semidefinite relaxation of Grothendieck's inequality.
\end{remark}

The two reductions rely on boxes, whose vertices encode the cuts of a graph. For two tokens in
centred balls with fixed values, as in the reductions, the largest score is the spectral norm
$\|Q\|$ scaled by the two radii, so the worst case is computed in polynomial time. The complexity
of certification over balls with more keys or with varying values is not settled here.

Both hardness results need the query to vary together with a key. When the query token is
fixed, as it is for a head whose reading position is not perturbed, every score depends on
one token only, and the problem becomes separable. For each varying key $u$, the pair
formed by its score and its value along the specification ranges over the planar set
\begin{equation}\label{eq:Cu}
  C_u=\bigl\{(s_u(z),h_u(z)):\ z\in\Z_u\bigr\}\subset\R^{2},
\end{equation}
while for a fixed key $k$ the pair $(s_k,h_k)$ is a constant. Let
$R=\max_j\hmax_j-\min_j\hmin_j$.

\begin{theorem}[Exact worst case with a fixed query]\label{thm:exactfq}
Let the query be fixed, and let the keys $u\in K_{\mathrm v}$ vary independently in compact
sets $\Z_u$ while the keys $k\in K_{\mathrm f}$, which include the query position, are fixed.
Then, the following hold.
\begin{enumerate}[label=\normalfont(\roman*), leftmargin=*]
\item The function
\begin{equation}\label{eq:H}
  H(\lambda)=\sum_{k\in K_{\mathrm f}}e^{s_k}\bigl(h_k-\lambda\bigr)+\sum_{u\in K_{\mathrm v}}\mu_u(\lambda),
  \qquad
  \mu_u(\lambda)=\max_{(s,h)\in C_u}e^{s}(h-\lambda),
\end{equation}
is convex and strictly decreasing from $+\infty$ to $-\infty$, its unique root is $\pstar$,
and tokens attaining the maxima $\mu_u(\pstar)$ attain the worst case.
\item If each $\Z_u$ is a box or zonotope with $r_u$ generators, or the convex hull of $m_u$ points,
then $C_u$ is a polygon with $V_u\le2r_u$ or $V_u\le m_u$ vertices, respectively. It is constructed from the projected generators or points in $O(r_u\log r_u)$ or $O(m_u\log m_u)$ operations, respectively. After this construction each evaluation of $H$ costs $O(|K_{\mathrm f}|+\sum_uV_u)$ operations, and $\pstar$ is computed to accuracy $\varepsilon$ with
$O(\log(R/\varepsilon))$ evaluations.
\item If each $\Z_u$ is an ellipsoid, then $C_u$ is a filled ellipse, possibly degenerate. The
maximum $\mu_u(\lambda)$ is then the largest value of $e^{s}(h-\lambda)$ at no more than five points
of the boundary of $C_u$, given by the real roots of a polynomial of degree at most four and one
further point. Each
evaluation of $H$ then costs $O(|K_{\mathrm f}|+|K_{\mathrm v}|)$ operations, and $\pstar$ is
computed to accuracy $\varepsilon$ with $O(\log(R/\varepsilon))$ evaluations.
\end{enumerate}
\end{theorem}

This is Dinkelbach's parametric approach to fractional
programming~\cite{dinkelbach1967nonlinear} made exact by separability. The output of the head is a ratio of two sums, and asking whether it can exceed $\lambda$ is asking whether a sum of
independent terms can be made nonnegative, which decouples over the keys. The theorem gives
the exact worst case together with an input that attains it, so it serves at once as an exact
certificate and as an exact attack. Dinkelbach's iteration, Newton's method on $H$ started
from the output of any input, produces outputs that increase monotonically and superlinearly to $\pstar$~\cite{schaible1976fractional}. Values of $\lambda$ with $H(\lambda)<0$ certify upper bounds. In the polygonal case $\mu_u(\lambda)$ is attained at a vertex of $C_u$ or at a
stationary point, in closed form, of the concave function $s+\log(h-\lambda)$ on an edge. For an ellipse the stationary points on the boundary are the roots of a polynomial of degree at most four. The cost is therefore linear in the number of keys up to logarithmic factors. On GPT-2, where each head reads its tokens through a subspace of dimension $2d_k+d_v=192$, in
which the token sets are ellipsoids, the theorem provides exact worst cases against which the convex certificates are measured at every layer. Discreteness is not a source of
hardness either.

\begin{remark}[Listed substitutions]\label{rem:subs}
Let each token be restricted to an explicit list of $m_j$ candidates, such as the synonyms of
a word. The worst case is the maximum, over the $m_i$ candidates of the query, of fixed query problems with finite key sets. Each of them is solved as the root of the separable equation of Theorem~\ref{thm:exactfq}(i), with $\mu_u$ a maximum of $m_u$ affine functions of $\lambda$. Once the $m_i\sum_um_u$ candidate scores have
been formed, $\pstar$ is computed to accuracy $\varepsilon$ in $O(m_i\sum_um_u\log(R/\varepsilon))$
operations, although there are $\prod_jm_j$ admissible inputs. The NP-hardness of Theorem~\ref{thm:hard} thus comes from the bilinear coupling of a varying query with a key over
sets described implicitly, such as boxes, and not from discreteness.
\end{remark}

\begin{table}[t]
\caption{Complexity of certifying one attention head. ``Exact'' means that the worst case is
computed to any prescribed accuracy with certified bounds.}
\label{tab:complexity}
\begin{tabular*}{\textwidth}{@{\extracolsep\fill}lll}
\toprule
setting & status & result\\
\midrule
scores in a box, values fixed & exact, $O(n\log n)$ & Remark~\ref{rem:threshold}\\
fixed query, boxes, zonotopes, point hulls & exact, polynomial & Thm.~\ref{thm:exactfq}(ii)\\
fixed query, ellipsoids & exact, polynomial & Thm.~\ref{thm:exactfq}(iii)\\
listed candidates, any query & exact, polynomial & Remark~\ref{rem:subs}\\
varying query, two tokens in boxes & NP-hard & Thm.~\ref{thm:hard}\\
same, score range in $[1.12,2]$ & inapproximable within $\frac{1}{200}$ of the range & Thm.~\ref{thm:inapprox}\\
\botrule
\end{tabular*}
\end{table}

Table~\ref{tab:complexity} summarizes the map. Convex certificates occupy the remaining territory, a varying query over convex sets. They are polynomial in every case, and they can be exact up to second order when the softmax slack weighted by the values is small. The inapproximability of Theorem~\ref{thm:inapprox} shows that, over boxes, nothing efficient can be uniformly better when this slack is not small.

\FloatBarrier

\section{Certificates}\label{sec:firstorder}

This section constructs two certificates. The first replaces the dependence of the attention pattern on the tokens by the set of patterns that the score intervals allow, and it lets the values move independently within their sets. It is a linear program solved greedily, with a first-order gap whose constant does not depend explicitly on the sequence length or the dimension. Its only source of error is that it decouples where a head attends from what
it reads. The second removes this decoupling by lifting. It is a level-one moment relaxation in which the softmax is represented by exponential cones and secants, and its gap is bounded by the value-weighted secant slack and is therefore of second order.

The pattern of attention is controlled by a \emph{certified pair} $(\hat a,\epsilon)$, a
reference pattern and a radius with $\|\sigma(s(z))-\hat a\|_1\le\epsilon$ for every admissible
input. The \emph{width pair} takes the midpoint pattern $\hat a=\sigma(s^{\mathrm{mid}})$ and
$\epsilon=\Delta/2$, which is certified by the Lipschitz bound of Lemma~\ref{lem:lip} because each score lies within
$\Delta/2$ of its midpoint. The \emph{margin pair}, available when $\gamma>0$, takes the
one-hot pattern of the dominant key and the radius of the margin bound~\eqref{eq:margin}.

\begin{definition}[Pattern polytope certificate]\label{def:pp}
For a certified pair $(\hat a,\epsilon)$, the \emph{pattern polytope} is
\begin{equation}\label{eq:polytope}
  \mathcal A=\Bigl\{a\in\simplex:\ \alo\le a\le\ahi,\ \|a-\hat a\|_1\le\epsilon\Bigr\},
\end{equation}
and the \emph{pattern polytope certificate} is
\begin{equation}\label{eq:rpp}
  \Rpp=\max_{a\in\mathcal A}\ \sum_{j\in J}a_j\,\hmax_j .
\end{equation}
\end{definition}

The certificate is valid. For every admissible input $z$, the pattern $a(z)$ lies in $\mathcal A$ and
\begin{equation*}
  \varphi(z)=\sum_ja_j(z)\,h_j(z_j)\le\sum_ja_j(z)\,\hmax_j .
\end{equation*} It is also the value of the convex relaxation of the head certification problem~\eqref{eq:pstar} in which each product $a_jz_j$ becomes a
variable constrained to the scaled set $a_j\Z_j$. It is computed by a two-phase transfer of
attention mass. Let $b$ be the projection of $\hat a$ onto the box $[\alo,\ahi]$ and
$\epsilon'=\epsilon-\|b-\hat a\|_1$ the remaining budget. The first phase restores $\sum_ja_j=1$. It adds the missing mass to the keys with the largest $\hmax_j$, or removes the excess from those with the smallest, each within its interval, at a cost of one unit of budget per unit of mass. The second phase transfers mass from the keys with the smallest $\hmax_j$ to those with the largest, within the same intervals and at a cost of two units of budget per unit of mass. It stops when the budget is exhausted or no transfer increases the objective.

\begin{proposition}[Greedy solution]\label{prop:greedy}
The two-phase transfer solves the linear program~\eqref{eq:rpp} exactly, so $\Rpp$ is computed
in $O(n\log n)$ time from the support values $\hmax_j$.
\end{proposition}

The proof (Appendix~\ref{app:proofs}) writes the transfer as two fractional knapsacks, one filling the keys with the largest support values and one emptying those with the smallest. The support values are in closed form for balls, ellipsoids and boxes. For the width pair the first phase is void, because $\hat a$ lies in the box and on the simplex. The algorithm then transfers a mass of at most $\epsilon/2$ from the keys with the lowest support values to those with the highest. The greedy algorithm makes the first-order certificate applicable to arbitrarily long contexts. Once the score intervals and the support values are known, it solves the linear program for $10^{6}$ keys in $1.4$ seconds. The accuracy of the certificate
is controlled by a gap bound of first order.

\begin{theorem}[Gap of the first-order certificate]\label{thm:B}
For every certified pair,
\begin{equation}\label{eq:thmB}
  0\le\Rpp-\pstar\le2\,\epsilon\,\radc ,
\end{equation}
which for the width pair reads $\Rpp-\pstar\le\Delta\,\radc$. If a key has margin $\gamma>0$,
the certificate with the width pair also satisfies
\begin{equation}\label{eq:thmC}
  \Rpp-\pstar\le\frac{4(n-1)e^{-\gamma}}{1+(n-1)e^{-\gamma}}\,\radc .
\end{equation}
\end{theorem}

The gap bound~\eqref{eq:thmB} is independent of the sequence length, the token dimension and the
head dimension, which enter only through the score width and the directional value radius. It
is of first order, because $\Delta$ is proportional to the perturbation radius. And it is \emph{directional}, since it involves the spread $\radc$ of the values along the specification rather than their spread in norm. On the trained heads of GPT-2, the latter is larger by a median factor $2.4$. In the margin regime the gap is exponentially small in the margin, so heads that attend decisively to one key are certified essentially exactly. The margin gap bound~\eqref{eq:thmC} holds for the width pair, without the cut of the margin pair, because the attention intervals imply that cut. The order of the theorem cannot be improved.

\begin{proposition}[Sharpness]\label{prop:sharp}
For every $\lambda\in(0,\tfrac12]$ and $\rho\le\tfrac12$ there is a two-key instance with a fixed
query and one token varying on a segment of half-length $\rho$, with score width
$\Delta=2\lambda\rho$ and directional value radius $\radc=\tfrac12(1+2\rho)$, on which
\begin{equation}\label{eq:sharp}
  \Rpp-\pstar=(1+2\rho)\tanh\Bigl(\frac{\lambda\rho}{2}\Bigr)=\tfrac12\,\Delta\,\radc\,\bigl(1+O(\lambda^{2}\rho^{2})\bigr).
\end{equation}
\end{proposition}

The first-order gap is thus of exact order $\rho$ while the values have a spread of order one,
and the first-order gap bound of Theorem~\ref{thm:B} is sharp up to a factor two. In the instance, which is written out in Appendix~\ref{app:proofs}, one key moves along a segment that raises its score while it lowers its value. The worst case pairs the most favourable value of the key with the attention that this value actually induces, which is the least attention. The pattern polytope instead pairs the most favourable value with the most favourable attention, a combination that no input realizes.

No bound computed from the score intervals and the ranges of the values along $c$ alone can do
better on this instance. If the key varies instead in the box $[-\rho,\rho]^{2}$, with its score read
from one coordinate and its value from the other, the score interval and the range of the value stay
the same. The worst case then puts the score at its upper end and the value at its maximum, and it
equals $\Rpp$. A bound that uses only this information and is valid for every instance must hold for both, so on
the segment its gap is at least that of $\Rpp$. This applies in particular to the maximum of
Remark~\ref{rem:threshold} over a box of scores with each value at its largest, which Rezazadeh and
Gholami Davoodi show to be the tightest bound that score intervals allow for fixed values~\cite{rezazadeh2026vertex}.
A certificate of higher order must therefore couple the scores with the values.

The greedy solution also explains the certificate. Its pattern $a^\star$ is a worst admissible
attention pattern, and the sensitivities of the linear program~\eqref{eq:rpp} to its data show which
keys and which quantities the bound depends on. Call the last key to receive mass in the transfer
the \emph{last receiver} $j^{+}$, and the last key to give mass the \emph{last donor} $j^{-}$.

\begin{proposition}[Sensitivity of the certificate]\label{prop:dual}
For the width pair, suppose that the greedy solution $a^\star$ is the unique optimal solution of the
linear program~\eqref{eq:rpp}, that it exhausts the budget, and that $j^{+}$ and $j^{-}$ lie strictly
inside their attention intervals. As a function of the data $\hmax$, $\epsilon$ and $\ahi$ of the
linear program, $\Rpp$ then has the following derivatives.
\begin{enumerate}[label=\normalfont(\roman*), leftmargin=*]
\item For every key $j$, $\partial\Rpp/\partial\hmax_j=a^\star_j$.
\item With respect to the radius of the pair, $\partial\Rpp/\partial\epsilon=\tfrac12\bigl(\hmax_{j^{+}}-\hmax_{j^{-}}\bigr)$.
\item For every receiver $j$ at its upper bound, $\partial\Rpp/\partial\ahi_j=\hmax_j-\hmax_{j^{+}}$.
\end{enumerate}
\end{proposition}

The certificate thus reads as a worst admissible pattern, the value that each key contributes along
the specification, the marginal value of the stability of the pattern, and the keys whose attention
bounds are binding.

To couple where a head attends with what it reads, a relaxation must carry the products of
attention weights with token perturbations as variables and relate them to the perturbations
themselves. Collect the base coordinates of all tokens and the attention weights in one vector,
$\xi=(u_1,\dots,u_n,a_1,\dots,a_n)$, where $u_i$ is the base coordinate of the query. Let $X$ be the moment matrix~\eqref{eq:moment} of $\xi$, which collects its first and second moments, with first moments $\bar u_k,\bar a_j$ and blocks
$Y[\cdot,\cdot]$. In the moment matrix, the score of key $j$ and the output along the
specification are, respectively, the linear functions
\begin{equation}\label{eq:liftedscore}
  \bar s_j(X)=\bar z_i^\top M\bar z_j+\bar z_j^\top M^\top B_i\bar u_i+\bar z_i^\top MB_j\bar u_j+\bigl\langle B_i^\top MB_j,\,Y[u_i,u_j]\bigr\rangle
\end{equation}
and
\begin{equation}\label{eq:liftedobj}
  \Phi(X)=\sum_{j\in J}\Bigl(\bar a_j\,h_j(\bar z_j)+g_j^\top Y[a_j,u_j]\Bigr),
\end{equation}
in which $Y[a_j,u_j]$ stands for the product of the attention weight of key $j$ with its
perturbation.

\begin{definition}[Level-one exponential cone relaxation]\label{def:R1}
For balls and ellipsoids in base coordinates, the relaxation $\Rlift$ is the conic program
\begin{alignat*}{2}
  \Rexp={}&\max_{X,\,\tau,\,t}\quad&&\Phi(X)\\
  &\text{s.t.}&&\|\bar u_k\|\le\rho_k,\quad\tr Y[u_k,u_k]\le\rho_k^{2},\tag{R1}\\
  &&&\slo_j\le\bar s_j(X)\le\shi_j,\quad t_j=\bar s_j(X)-\tau,\quad\tlo_j\le t_j\le\thi_j,\quad\taulo\le\tau\le\tauhi,\tag{R2}\\
  &&&(\bar a_j,1,t_j)\in\Kexp,\quad\bar a_j\le\ell_j(t_j),\quad\alo_j\le\bar a_j\le\ahi_j,\quad\textstyle\sum_{j\in J}\bar a_j=1,\tag{R3}\\
  &&&Y[a_j,a_j]\le(\alo_j+\ahi_j)\bar a_j-\alo_j\ahi_j,\tag{R4}\\
  &&&\begin{aligned}
    &\|Y[a_j,u_j]\|\le\rho_j\bar a_j,\quad\|\bar u_j-Y[a_j,u_j]\|\le\rho_j(1-\bar a_j),\\
    &\textstyle\sum_{j\in J}Y[a_j,u_k]=\bar u_k,\quad\sum_{j\in J}Y[a_j,a_l]=\bar a_l,\quad Y[a_j,a_l]\ge0,
  \end{aligned}\tag{R5}\\
  &&&\|\bar a-\hat a\|_1\le\epsilon,\tag{R6}\\
  &&&X\in\Kpsd,\quad X_{00}=1,\tag{R7}
\end{alignat*}
where $k$, $j$ and $l$ range over $J$, and $(\hat a,\epsilon)$ is a certified pair.
\end{definition}

The seven groups of constraints have the following roles. The \emph{lifted sets} (R1) keep the first
moments in the balls and bound the second moments by the trace that the rank-one moment matrix of an
input attains. The \emph{scores and log-attention} (R2) keep the lifted scores in their certified
intervals and define the log-attention through the normalizer $\tau$. The \emph{softmax sandwich} (R3)
imposes $\bar a_j\ge e^{t_j}$ exactly through the exponential cone, bounds $\bar a_j$ from above by the
secant $\ell_j$, and keeps the weights in their intervals and on the simplex. The \emph{interval
diagonal} (R4) is the product of the two interval constraints of a weight,
$(a_j-\alo_j)(\ahi_j-a_j)\ge0$. It bounds the variance of the lifted weight, which second-order accuracy requires.
In computations, replacing it by the weaker $Y[a_j,a_j]\le\bar a_j$ leaves gaps of first order. The \emph{products
with the sets} (R5) are products of the simplex and of the set constraints with each other. They do
not enter the accuracy bound, but they tighten the relaxation and imply $\Rexp\le\Rpp$ for the same
certified pair. They bound the product of each weight with its perturbation by the weight times the
radius, so the lifted objective never exceeds the objective of the pattern polytope at the first
moments $\bar a$ (Appendix~\ref{app:proofs}). The \emph{pattern cut} (R6) restricts the lifted pattern
in the same way as the pattern polytope~\eqref{eq:polytope}. The \emph{moment conditions} (R7) require $X$ to be positive semidefinite with a unit constant entry, as the moment matrix of every input is. The moment matrix of an input also has rank one, which is not imposed because it is not a convex constraint.

The relaxation is valid, because for every admissible input the rank-one moment matrix of $(u,a(z))$,
together with $\tau=\lse(s(z))$ and $t_j=\log a_j(z)$, satisfies all constraints. Boxes are treated
analogously (Appendix~\ref{app:relaxation}). The relaxation has one positive semidefinite cone of
side $1+\sum_kr_k+n$, $n$ exponential cones and a number of second-order cones linear in $n$.
The main result of the paper bounds its gap by the value-weighted secant slack, plus two terms
that account for the second moments.

\begin{theorem}[Second-order accuracy]\label{thm:F0v}
For every instance with ball, ellipsoid or box token sets,
\begin{equation}\label{eq:F0v}
  0\le\Rexp-\pstar\le\KN(\chi,D,D_{\mathrm{tot}})+\eta\,\radc+\frac12\sum_{j\in J}\bigl(\ahi_j-\alo_j\bigr)\nu_{c,j},
\end{equation}
where, for boxes, $\eta$ is computed from the circumradii.
\end{theorem}

The three terms correspond to the three ways in which a moment matrix can differ from the moment matrix of an input. The proof (Appendix~\ref{app:proofs}) splits the objective of a feasible moment matrix at its first moments, which form an admissible input, and bounds each of the three deviations in turn. The first is the slack of the softmax sandwich, weighted as in the value-weighted sandwich identity of Proposition~\ref{prop:P7v}. The relaxation can inflate the attention of key $j$ by at most
$D_j$, and this helps only if the value of key $j$ can exceed the output. The second is the covariance between the query and key perturbations. It shifts the lifted scores by at most $\eta$ and hence, by the Lipschitz bound of Lemma~\ref{lem:lip}, the pattern by at most $\eta$ in $\ell_1$ norm. The
third is the covariance between an attention weight and the perturbation of its own token, which
the interval diagonal bounds by the product of the attention uncertainty and the directional value width.
Every term is computable before the relaxation is solved, so the bound can decide whether the
semidefinite program is worth solving and where to branch.

The value weighting is what makes the bound informative on trained heads. Bounding the total
variation of the relaxed pattern instead of its value-weighted effect leads to the same structure, with the first term replaced by $2\radc$ times a measure of the total slack such as $2S/(1-S)$. That measure blows up as $S\to1$ and charges every key at the full range of the values. With
$\phimin\ge\min_l\hmin_l$, which may be assumed because $\Rexp$ does not depend on $\phimin$, the first term of the second-order bound~\eqref{eq:F0v} is at most $(2+\eta)\radc\min\{S,D_{\mathrm{tot}}\}$. This bound vanishes quadratically in the score width, and the knapsack term itself can be far smaller, because keys whose values cannot exceed $\phimin-\eta\,\radc$ carry no weight. To state the orders explicitly, scale the radii as $\rho_k=\rho\,\hat\rho_k$ with $\hat\rho_k\le1$ and $\|B_k\|\le1$. Let
$R_0=\max_k\|\bar z_k\|$, $\beta_c=\|N^\top c\|$ and $r_{0,c}$ the half-range of the nominal values
$h_j(\bar z_j)$, let $\rho_1\le1$ be the largest radius with $2\|M\|(2R_0\rho_1+\rho_1^{2})\le1$, and
define
\begin{equation}\label{eq:Kh}
\begin{aligned}
  K_h&=(r_{0,c}+\beta_c)\bigl(5e\,\|M\|^{2}(2R_0+1)^{2}+\|M\|\bigr)+2.4\,\beta_c\,\|M\|(2R_0+1),\\
  K'_h&=2\|M\|(2R_0+1)(r_{0,c}+\beta_c).
\end{aligned}
\end{equation}
For a certificate $r(\rho)$ and a tolerance $\varepsilon>0$, the \emph{tightness radius}
$\bar\rho(\varepsilon)$ is the largest $\bar\rho$ such that $r(\rho)-\pstar(\rho)\le\varepsilon$ for
all $\rho\le\bar\rho$.

\begin{corollary}[Orders of accuracy]\label{cor:orders}
For every instance with ball, ellipsoid or box token sets, with $\eta$ computed from the circumradii for boxes, the following hold.
\begin{enumerate}[label=\normalfont(\roman*), leftmargin=*]
\item The gap of the second-order certificate satisfies
\begin{equation}\label{eq:orderS}
\begin{aligned}
  \Rexp-\pstar&\le\radc\Bigl((2+\eta)\min\{S,D_{\mathrm{tot}}\}+\eta\Bigr)+\tfrac12\nu_cW\\
  &\le\radc\Bigl((2+\eta)\tfrac{\Delta^{2}}{2}e^{\Delta}+\eta\Bigr)+\nu_c\sinh\Delta .
\end{aligned}
\end{equation}
\item If the radii are $\rho_k=\rho\,\hat\rho_k$ with $\hat\rho_k\le1$ and $\|B_k\|\le1$, then
$\Rexp-\pstar\le K_h\rho^{2}$ and $\Rpp-\pstar\le K'_h\rho$ for $\rho\le\rho_1$, with the constants
$K_h$ and $K'_h$ defined in~\eqref{eq:Kh}, so the tightness radii satisfy
\begin{equation}\label{eq:radii}
  \bar\rho^{\exp}(\varepsilon)\ge\min\Bigl\{\rho_1,\sqrt{\varepsilon/K_h}\Bigr\},
  \qquad
  \bar\rho^{\mathrm{pp}}(\varepsilon)\ge\min\Bigl\{\rho_1,\varepsilon/K'_h\Bigr\},
\end{equation}
and on the two-key instances of Proposition~\ref{prop:sharp} the first-order tightness radius is
$O(\varepsilon)$.
\end{enumerate}
\end{corollary}

Second-order certification thus halves the exponent of the tightness radius in the tolerance, so halving the tolerance shrinks the certifiable radius by a factor $\sqrt2$ instead of $2$. The constants follow from the order bound~\eqref{eq:orderS} once the score intervals are bounded by the Cauchy--Schwarz inequality (Appendix~\ref{app:proofs}). They depend on the head only through $\|M\|$, $\|N^\top c\|$, the largest nominal token norm and the nominal spread of the values along $c$. The length of the context and the dimension of the tokens therefore enter only through these norms. For layer-normalized tokens, for instance, the largest nominal norm grows like the square root of the dimension times the gain of the normalization. The accuracy also depends on the temperature of the head, through $\|M\|$, and on its gain along the specification. As a function of the weights, $K_h$ is locally Lipschitz and differentiable almost everywhere, which makes it a natural regularizer for training heads that are certifiable at a prescribed radius. The tightness radius measures the accuracy of the certificate, not the sensitivity of the model. The largest radius at which the certified worst
case stays within $\varepsilon$ of the nominal output is of first order for every certificate,
because the output itself moves by $\Theta(\rho)$.

The order two of these bounds cannot be improved. The sharpness instance of
Proposition~\ref{prop:sharp} does not show this. Its worst case puts the score at an end of its
interval, where the secant slack vanishes, and solving $\Rlift$ on it returns $\pstar$ to solver
precision. Letting the key move in a disc separates the second-order certificate from the truth at
second order.

\begin{proposition}[The second order is attained]\label{prop:lower}
For every $\lambda>0$ there is a two-key instance with a fixed query and one token varying in a
disc of radius $\rho$, with score width $\Delta=2\lambda\rho$, on which
\begin{equation}\label{eq:lower}
  \liminf_{\rho\to0}\frac{\Rexp-\pstar}{\rho^{2}}\ge\frac{\lambda^{2}}{32+2\lambda^{2}} .
\end{equation}
On this instance, with $\phimin$ from the first-order certificate for $-c$, the value-weighted
secant slack equals $\tfrac{1}{32}\lambda^{2}\rho^{2}\bigl(1+o(1)\bigr)$.
\end{proposition}

The score of the key moves along one axis of the disc and its value along the other. The proof
(Appendix~\ref{app:proofs}) takes a worst-case input and raises the attention of the key, whose
value exceeds the output, by the slack that the secant leaves at its score. The moment matrix of
this input, with the raised attention in place of the softmax, is feasible for $\Rlift$. The first
term of the second-order bound~\eqref{eq:F0v} is thus attained up to the factor
$1+\lambda^{2}/16$. The third term, $\tfrac12(\ahi_1-\alo_1)\nu_{c,1}\approx\tfrac12\lambda\rho^{2}$, bounds a covariance between the weight and the perturbation of the key that this rank-one moment matrix does not have. The tightness radius of the second-order certificate is of exact order
$\sqrt{\varepsilon}$ on this instance. Solving $\Rlift$ with $\lambda=\tfrac12$, with Clarabel~\cite{goulart2024clarabel} through CVXPY~\cite{diamond2016cvxpy} at a tolerance of $10^{-9}$, gives
$(\Rexp-\pstar)/\rho^{2}$ between $0.0080$ and $0.0122$ for $\rho$ from $0.0125$ to $0.2$, against
the limit $1/130\approx0.0077$ of the lower bound~\eqref{eq:lower}.

The attained second order of Proposition~\ref{prop:lower} does not contradict the local stability
of semidefinite relaxations. Cifuentes, Agarwal, Parrilo and Thomas~\cite{cifuentes2022local}
consider parametric families of quadratically constrained quadratic programs whose semidefinite
relaxation is exact at a nominal parameter. Under regularity conditions, they show that the
relaxation remains exact in a neighbourhood of that parameter. Here, the nominal problem, with
$\rho=0$, is solved exactly by the relaxation, since every token set is a point, yet on the instance
of Proposition~\ref{prop:lower} the gap is of exact order $\rho^{2}$ for all small $\rho>0$. Their
regularity conditions fail at $\rho=0$, and enlarging the sets from a point produces gaps of second
order even for quadratic problems. By homogeneity, for instance, the Shor relaxation of
\begin{equation*}
  \max\bigl\{u^\top Qu':\ \|u\|_\infty\le\rho,\ \|u'\|_\infty\le\rho\bigr\}
\end{equation*}
has the gap $\rho^{2}\bigl(\mathrm{SDP}(Q)-\|Q\|_{\infty\to1}\bigr)$, where $\mathrm{SDP}(Q)$ is the
value of the relaxation at $\rho=1$. For attention, the gap is created by the secant representation
of the exponential, since the feasible point in the proof of Proposition~\ref{prop:lower} has a
moment matrix of rank one. The second order of Theorem~\ref{thm:F0v} is thus the exact order of the
level-one relaxation, not an artefact of its analysis.

The first-order gap bound of Theorem~\ref{thm:B} and the second-order bound of Theorem~\ref{thm:F0v} delimit three regimes. When the score width is small, so
are the value-weighted secant slack, the score covariance and the attention uncertainty, and the
second-order certificate is accurate to second order. When one key dominates with a margin
$\gamma$, the first-order certificate is exact up to a term exponentially small in $\gamma$,
whatever the score width. When the query varies and the score width is of order one with several keys competing, the slack is of order one. For tokens in boxes, the inapproximability result of Theorem~\ref{thm:inapprox} shows that no efficient method can then do uniformly better. In this regime, which on synthetic heads appears as a band of intermediate widths, all methods are loose, and branching over the scores is the remedy. For a deployed model this gives a design rule. \emph{A head is certified to accuracy $\varepsilon$ when its attention logits cannot swing by more than a fixed fraction of $\sqrt{\varepsilon/\radc}$ under the admissible perturbations and the remaining terms of the order bound~\eqref{eq:orderS} together stay below $\varepsilon/2$. It is also certified when one key dominates by a margin $\gamma\ge\log\bigl(4(n-1)\radc/\varepsilon\bigr)$.}

\section{Implementation}\label{sec:computation}

A certificate is only as good as its computation. Interior-point solvers return approximate
primal and dual solutions in floating point, and a value that is slightly too small is not a
certificate at all. The token dimension of a trained model is large, and the score intervals
that feed every bound must themselves be certified. This section resolves each of these
issues and describes how branching over the scores tightens the certificate where it is loose. Algorithm~\ref{alg:cert} summarizes the procedure.

A head reads each token only through its queries, keys and values. With biases, the scores and
values are affine functions of $P_hz$, where $P_h=[W_Q\ W_K\ W_V]^\top\in\R^{d'\times d}$ and $d'=2d_k+d_v$. For GPT-2 small, $d'=192$ against $d=768$. Certificates can be computed in coordinates of dimension at most $d'$ without any loss.

\begin{proposition}[Exact reduction]\label{prop:projection}
Let the token sets be balls in $\R^{d}$ and let the columns of $V$ form an orthonormal basis of the row space of $P_h$. Replacing each base coordinate $u$ by $V^\top u$ changes neither
$\pstar$, nor $\Rpp$, nor the value of $\Rlift$, nor the semidefinite score intervals.
\end{proposition}

The reason is that the scores, the values along $c$ and every lifted expression depend on a base coordinate $u$ only through its projection $V^\top u$ (Appendix~\ref{app:proofs}). In the reduced coordinates each ball becomes
\begin{equation*}
    P_h\Z_j=\bigl\{P_h\bar z_j+P_hVv:\ \|v\|\le\rho_j\bigr\},
\end{equation*}
with the biases absorbed into the centres. For GPT-2 the reduction shrinks the semidefinite
variable of a head at query position $i$ from side $1+768\,(i+1)+(i+1)$ to
$1+192\,(i+1)+(i+1)$, by a factor of about sixteen in the number of entries. It is what makes the second-order certificate affordable for trained models.

Every bound depends on the certified score intervals~\eqref{eq:scoreint}. The score of key $j$ is
a constant plus a linear function of the query and key perturbations plus the bilinear term
$u_i^\top Qu_j$, $Q=B_i^\top MB_j$. Its extreme values over two balls are bounded by the Shor relaxation, and the endpoints of the resulting interval are widened by a relative $10^{-7}$ to cover the solver's tolerance.
The bound is exact for the self-score of the query, a quadratic function of one ball, by the S-lemma. It is also exact for the bilinear term over two centred balls, with value
$\rho_i\rho_j\|Q\|$, because the off-diagonal block of a positive semidefinite matrix whose
diagonal blocks have traces at most $\rho_i^{2}$ and $\rho_j^{2}$ has nuclear norm at most $\rho_i\rho_j$. Over two centred boxes it is within the factor $K_G$ of the maximum~\cite{alon2006approximating}. With the linear terms present the problem is a generalized
two-ball trust-region problem, for which the bound can be loose. On offset synthetic instances its ratio to the exact maximum reached $1.40$ on balls and $1.23$ on boxes. At the radii used for
GPT-2 the linear terms dominate, and the semidefinite intervals are a median factor $3.3$ narrower
than those of the Cauchy--Schwarz inequality. Each interval is computed in coordinates of dimension
$2d_k$, by the argument of the exact reduction of Proposition~\ref{prop:projection}, in a fraction of a second per key (Appendix~\ref{app:relaxation}).

An interior-point solver returns an approximate dual solution whose objective may lie slightly
below the optimal value of the relaxation. Taken at face value, such a number can fall below the true worst case. In a preliminary run on synthetic heads, the solver's objective fell below
a feasible value on $230$ of $60{,}154$ semidefinite instances, by up to $6.6\times10^{-4}$. The
remedy is a bound that is valid for any dual point. Write the relaxation as
\begin{equation}\label{eq:generic}
  \max\Bigl\{\langle F,X\rangle+f^\top x+f_0:\ \mathcal A_k(X,x)+b_k\in\mathcal K_k\ (k=1,\dots,K),\ X\in\Kpsd\Bigr\},
\end{equation}
with linear maps $\mathcal A_k$ and cones $\mathcal K_k$. Suppose that every feasible point
satisfies a priori bounds $x^{\mathrm{lo}}\le x\le x^{\mathrm{hi}}$ and $X_{ll}\le\delta_l$, together
with group budgets $\sum_{l\in G_q}X_{ll}/\delta_l\le1$. For $\Rlift$ these follow from the interval
constraints, the trace constraints of the balls and the interval diagonal. For multipliers $\lambda_k$, let
\begin{equation}\label{eq:residuals}
    f_\lambda=f+\sum_k\mathcal A_k^{x*}\lambda_k,
  \qquad
  F_\lambda=F+\sum_k\mathcal A_k^{X*}\lambda_k,
  \qquad
  f_{0,\lambda}=f_0+\sum_k\lambda_k^\top b_k,
\end{equation}
where $\mathcal A_k^{x*}$ and $\mathcal A_k^{X*}$ are the
adjoints of the parts of $\mathcal A_k$ acting on $x$ and $X$. Finally, let $n_G$ be the number of groups plus the number of ungrouped indices with $\delta_l>0$.

\begin{proposition}[Safe dual bound]\label{prop:safe}
For any multipliers $\lambda_k\in\mathcal K_k^{*}$, the optimal value of problem~\eqref{eq:generic}
is at most
\begin{equation}\label{eq:safe}
    f_{0,\lambda}+\sum_l\max\bigl\{(f_\lambda)_lx^{\mathrm{lo}}_l,\ (f_\lambda)_lx^{\mathrm{hi}}_l\bigr\}
  +n_G\,\lambda_{\max}^{+}\bigl(\diag(\delta)^{1/2}F_\lambda\diag(\delta)^{1/2}\bigr).
\end{equation}
\end{proposition}

The bound is weak duality with the residual of the dual constraints absorbed by the a priori
bounds, in the spirit of the rigorous error bounds of Jansson et al.\ and of Neumaier and
Shcherbina~\cite{jansson2007rigorous,neumaier2004safe}. The multipliers are the solver's dual solution projected onto the dual cones, and an explicit allowance covers the rounding errors of the evaluation (Appendix~\ref{app:relaxation}). The resulting \emph{safe value} replaces the solver's objective everywhere. Because it is valid for any dual point, the solver can be stopped at moderate tolerances. On GPT-2, feasibility tolerances of $10^{-6}$ reduce the solve time by a factor
five, and the safe value exceeds the solver's dual objective by a median $2\times10^{-7}$. Strict feasibility is not needed either. This matters because $\Rlift$ has no strictly feasible point, since the equalities in (R3), (R5) and (R7) give $Xv=0$ for the vector $v$ with entry $-1$ at the
constant, zeros at the token coordinates and ones at the attention weights. Eliminating one weight
through $\sum_ja_j=1$ removes this particular kernel and is advisable for solvers that rely on interior points.

\begin{definition}[Certificate]\label{def:cert}
The certificate of a head instance is
\begin{equation}\label{eq:cert}
  \cert=\min\bigl\{\Rpp,\ \text{safe value of }\Rlift\bigr\}.
\end{equation}
\end{definition}

Both members of the minimum are valid upper bounds. The optimal value of $\Rlift$ never exceeds
$\Rpp$, but its safe value can, which is one reason for the minimum. The others are that $\Rpp$ is
computed in $O(n\log n)$ time, is essentially exact in the margin regime, and serves as a fallback
when the semidefinite program is not solved.

When the certificate is loose, it can be tightened by partitioning the admissible set in score
space. Splitting the interval of key $j$ at its midpoint divides the inputs into those with $s_j(z)$ below and above the midpoint. Each part is certified by $\Rlift$ with the lifted
score~\eqref{eq:liftedscore} constrained to the half interval and the controls recomputed. The maximum over the leaves of a partition is a certificate. Leaves whose certificate does not exceed a known feasible value are discarded, and a leaf whose relaxation is infeasible contains no input. The second-order bound of Theorem~\ref{thm:F0v} applies to each leaf with the controls of the leaf and with the global worst case $\pstar$ in place of the leaf's own. Only $\phimin$ must remain a lower bound over the whole admissible set. The reason is that the first moments of a leaf's relaxation are admissible inputs of the
whole problem but not necessarily of the leaf. As the score intervals shrink, the first and third terms of the second-order bound~\eqref{eq:F0v} vanish, while $\eta\,\radc$ does not depend on them. Score branching therefore drives the gap to at most $\eta\,\radc$, which is zero for a fixed query. The bound also chooses the split. The \emph{bound-lookahead} rule computes the controls of both children of every
candidate key in closed form, evaluates the second-order bound~\eqref{eq:F0v} on each, and splits the key whose larger child bound is smallest. It costs $O(n^{2})$ closed form evaluations per node. Splitting
the key with the largest raw slack is often ineffective. The largest slack often sits on keys whose
values cannot exceed the output, and by the value-weighted sandwich identity of
Proposition~\ref{prop:P7v} the slack of such keys does not raise the bound.

\begin{algorithm}[t]
\caption{Certificate for one head, one query position and one specification}\label{alg:cert}
\begin{algorithmic}[1]
\Require weights of the head, token sets $\Z_j$, specification $c$, tolerance $\varepsilon$
\State reduce the base coordinates to the head subspace (Proposition~\ref{prop:projection})
\State compute the score intervals by the semidefinite bounds
\State compute the controls of Table~\ref{tab:controls}, with $\phimin$ from the greedy certificate for $-c$
\State $\Rpp\gets$ greedy solution of the pattern polytope linear program~\eqref{eq:rpp} (Proposition~\ref{prop:greedy})
\If{the margin gap bound~\eqref{eq:thmC} is at most $\varepsilon$}
  \State \Return $\Rpp$
\EndIf
\State solve $\Rlift$ by an interior-point method; compute its safe value (Proposition~\ref{prop:safe})
\State $\cert\gets\min\{\Rpp,\text{safe value}\}$
\While{$\cert$ minus the best feasible value exceeds $\varepsilon$ and the budget allows}
  \State split the open leaf with the largest bound by the bound-lookahead rule; certify both children
  \State $\cert\gets$ maximum of the open leaves' certificates; improve the feasible value from the leaves' first moments
\EndWhile
\State \Return $\cert$
\end{algorithmic}
\end{algorithm}

The cost is dominated by the semidefinite solves. For a head at query position $i$ the reduced
lifted vector has dimension $(i+1)(d'+1)$, that is $965$, $1{,}737$ and $2{,}509$ for $i=4$, $8$ and $12$, respectively, on GPT-2. With the task interface of MOSEK~\cite{mosek2026mosek}, the GPT-2 solves with a varying query take a median $22$, $115$ and $382$ seconds, respectively, on four threads, with memory below $18$ gigabytes. A modelling layer that dualizes the relaxation before
passing it to the solver turns every entry of the moment matrix into an equality constraint. A single instance at $i=8$ then exhausted $186$ gigabytes, so the direct formulation is essential. The greedy certificate and all controls cost milliseconds to seconds. The exact fixed query solver of
Theorem~\ref{thm:exactfq} costs a median $0.13$ seconds per GPT-2 head, prompt, query position and radius, and grows linearly with the
number of keys, from $0.11$ seconds at ten keys to $115$ seconds at ten thousand in dimension eight.

The single positive semidefinite cone grows with the context, with side $1+(i+1)(d'+1)$ at query
position $i$. The proof of the second-order bound of Theorem~\ref{thm:F0v} uses the moment matrix
only through its principal submatrices indexed by the constant, the query, one key and its weight
(Appendix~\ref{app:proofs}). Requiring only these $n$ submatrices to be positive semidefinite therefore gives a weaker
relaxation to which the second-order bound~\eqref{eq:F0v} still applies. It keeps (R1)--(R4), (R6)
and the norm constraints of (R5) and drops the sums and sign constraints of (R5). Its cones have side at most $2d'+2$, so
its size grows linearly with the number of keys. How much accuracy it gives up against $\Rlift$ is
not measured here.

\section{Per-Operation Relaxations}\label{sec:perop}

Linear bound propagation, its optimized variants and polyhedral abstract domains relax a network one operation at a time. Each nonlinear operation is replaced by affine bounds that are valid on the interval, or the box, of its operands, and the bounds are composed by back-substitution. The second-order certificate of Section~\ref{sec:firstorder}, a moment relaxation of the whole head, does the opposite. It couples the second-order terms of all keys through one moment matrix. This section shows what
the per-operation approach costs at second order. On a function whose value and gradient
vanish at the nominal point, a per-operation relaxation pays the sum of the operations'
individual second-order maxima, whereas the true second-order term is one coupled maximum.
This is a second-order counterpart of the convex barrier of Salman et
al.~\cite{salman2019convex}.

Such functions arise as second-order parts and remainders. Linearizing a transformer block $F$ at its nominal input $\bar x$ leaves the remainder
\begin{equation}\label{eq:blockrem}
  r(\delta)=F(\bar x+\delta)-F(\bar x)-DF(\bar x)\,\delta,
\end{equation}
which vanishes with its gradient at $\delta=0$, and so does every linear functional of it. Certifying a computation through several blocks requires bounds on such remainders, and bound propagation applied to the network that computes the remainder is the natural source of them. Consider a symmetric compact convex input set $\mathcal D$ and operations $o$ with operands affine
in the perturbation, $x_o(\delta)=\bar x_o+G_o\delta\in\R^{k_o}$. Given twice continuously
differentiable functions $\phi_o:\R^{k_o}\to\R$ and weights $\lambda_o\in\R$, the second-order part
of their weighted sum is
\begin{equation}\label{eq:psiT}
  \Psi(\delta)=\sum_o\lambda_o\,r_o\bigl(x_o(\delta)\bigr),
  \qquad
  r_o(x)=\phi_o(x)-\phi_o(\bar x_o)-\nabla\phi_o(\bar x_o)^\top(x-\bar x_o).
\end{equation}
Let $\omega_o\in\R^{k_o}$ collect the half-widths of the operands and let $\mathcal B_o$ be the operand box, namely
\begin{equation}\label{eq:opbox}
  \omega_{o,i}=\max_{\delta\in\mathcal D}(G_o\delta)_i,
  \qquad
    \mathcal B_o=\prod_i\bigl[\bar x_{o,i}-\omega_{o,i},\ \bar x_{o,i}+\omega_{o,i}\bigr].
\end{equation} A
\emph{per-operation relaxation bound} is a number
\begin{equation}\label{eq:Uper}
  U=\max_{\delta\in\mathcal D}\sum_o\ell_o\bigl(x_o(\delta)\bigr),
\end{equation}
in which each $\ell_o$ is affine and satisfies $\ell_o\ge\lambda_or_o$ on a box containing $\mathcal B_o$.
For one layer of operations, an affine upper bound of $\lambda_o\phi_o$ on the box differs from such an $\ell_o$ by an affine function. The bounds of linear bound propagation~\cite{zhang2018efficient,xu2020automatic}, of its optimized variant with any choice of
slopes~\cite{xu2021fast} and of DeepPoly-type domains~\cite{singh2019abstract} are therefore of the form~\eqref{eq:Uper}. Part~(iii) of the next proposition covers composed networks.

\begin{proposition}[Per-operation relaxations pay a sum of maxima]\label{prop:T}
The following three statements hold.
\begin{enumerate}[label=\normalfont(\roman*), leftmargin=*]
\item Every per-operation relaxation bound satisfies
\begin{equation}\label{eq:Tlower}
  U\ge\sum_o\max\Bigl\{0,\ \max_{\varsigma\in\{\pm1\}^{k_o}}\tfrac12\bigl[\lambda_or_o(\bar x_o+\varsigma\circ\omega_o)+\lambda_or_o(\bar x_o-\varsigma\circ\omega_o)\bigr]\Bigr\}.
\end{equation}
\item If $K_o\succeq\lambda_o\nabla^{2}\phi_o$ on $\mathcal B_o$ for every $o$, then
$\Psi(\delta)\le\frac12\delta^\top\mathcal Q\delta$, where $\mathcal Q=\sum_oG_o^\top K_oG_o$, with equality when
every $\phi_o$ is quadratic and $K_o=\lambda_o\nabla^{2}\phi_o$.
\item If $\Psi$, with $\Psi(0)=0$, is computed by a graph whose operands may depend nonlinearly
on $\delta$, and $U$ is obtained by back-substituting affine bounds valid on intervals
$I_o\supseteq x_o(\mathcal D)$, then the lower bound~\eqref{eq:Tlower} holds with the back-substitution
multipliers $\hat\lambda_o$ in place of $\lambda_o$ and any $\omega_o$ with
$\bar x_o\pm\varsigma\circ\omega_o\in I_o$ for all $\varsigma\in\{\pm1\}^{k_o}$.
\end{enumerate}
\end{proposition}

For a unary operation the term of the lower bound~\eqref{eq:Tlower} is at least
$\frac12(\underline\kappa_o)_+\omega_o^{2}$, where $\underline\kappa_o$ is the minimum of $\lambda_o\phi_o''$ on the operand interval. For a product $\phi_o(x,y)=xy$ it is $|\lambda_o|\omega_{o,x}\omega_{o,y}$. On a ball of radius $\rho$, part~(ii) gives
$\max\Psi\le\frac12\rho^{2}\lambda_{\max}(\mathcal Q)_+$. The mechanism of part~(i) is elementary. An affine function that dominates a remainder at two opposite corners of a box is, at the centre, at least the average of the remainder's two corner values, although the remainder vanishes there. The nominal input is the centre of every operand box at once. In a composed network the
multipliers $\hat\lambda_o$ depend on the slopes chosen downstream. If the slopes tend to the
derivatives at the nominal point as the radius tends to zero, then $\hat\lambda_o$ tends to the adjoint $\lambda_o=\partial\Psi/\partial(\text{output of }o)$ and the admissible half-widths to the linearized ones, $\rho\|g_o\|$, where $g_o$ is the gradient in $\delta$ of a scalar operand. The right side of the lower bound~\eqref{eq:Tlower} evaluated with these is then the leading-order lower bound for such relaxations. Over a ball, the comparison with the exact
term reduces to a comparison between a trace and an eigenvalue.

\begin{corollary}[The effective rank factor]\label{cor:T}
Let $\mathcal D$ be a ball of radius $\rho$ and let every operation be unary with constant curvature
$\lambda_o\phi_o''=\underline\kappa_o\ge0$ and row $g_o=G_o$, so that
$\mathcal Q=\sum_o\underline\kappa_og_og_o^\top$. If $\mathcal Q\ne0$, then
\begin{equation}\label{eq:corT}
  \max_{\mathcal D}\Psi=\tfrac12\rho^{2}\lambda_{\max}(\mathcal Q),
  \qquad
  U\ge\tfrac12\rho^{2}\tr\mathcal Q,
\end{equation}
and the level-one Shor relaxation of $\max_{\mathcal D}\Psi$ attains
$\frac12\rho^{2}\lambda_{\max}(\mathcal Q)$.
\end{corollary}

Every per-operation relaxation thus overestimates the second-order term by at least the factor
$\tr\mathcal Q/\lambda_{\max}(\mathcal Q)$, the effective rank, which lies between one and $\rank\mathcal Q$, while a moment relaxation does not. The variance $\|Py\|^{2}/d$ inside a layer normalization over a ball of radius $\rho$ is the simplest instance. Its exact second-order term is $\rho^{2}/d$, while a coordinate-wise relaxation pays $(d-1)\rho^{2}/d$, a factor $d-1=767$ in GPT-2
small. Two consequences follow. First, optimizing the slopes of linear bound propagation cannot remove the effect. For one layer the lower bound~\eqref{eq:Tlower} does not depend on the slopes, and for composed networks, under the condition above, its leading term does not. Second, the lifted relaxation of a single head applies the coupling of Corollary~\ref{cor:T}, an eigenvalue in place of a trace, at the scale of a head, which is consistent with its advantage over bound propagation on single heads. This is an interpretation rather than a consequence of the proposition, because the output of a head has a linear part, whereas the proposition concerns second-order parts and remainders.

The two sides can be evaluated on a trained model at the nominal point. For a function $\Psi$
computed by a graph of affine maps, unary operations and products, with $\Psi(0)=0$ and
$\nabla\Psi(0)=0$, let $\lambda_o$ be the adjoint of operation $o$ and $g_o$, or $g_{o,x},g_{o,y}$, the gradients of its operands in $\delta$, all at $\delta=0$, and let $\kappa_o=\phi_o''(\bar x_o)$. Write
$H^{+}$ for the positive semidefinite part of a symmetric matrix $H$.

\begin{lemma}[Spectral and per-operation second-order terms]\label{lem:chain}
The Hessian of $\Psi$ at the nominal point is
\begin{equation}\label{eq:H0}
  H_0=\sum_{o\ \mathrm{unary}}\lambda_o\kappa_og_og_o^\top+\sum_{o\ \mathrm{product}}\lambda_o\bigl(g_{o,x}g_{o,y}^\top+g_{o,y}g_{o,x}^\top\bigr),
\end{equation}
and on a ball of radius $\rho$, with $s_{\mathrm{spec}}=\frac12\rho^{2}\lambda_{\max}(H_0)_+$,
\begin{equation}\label{eq:chain}
  s_{\mathrm{spec}}\le\tfrac12\rho^{2}\tr(H_0^{+})\le s_{\mathrm{op}}
  \eqdef\sum_{o\ \mathrm{unary}}\tfrac12(\lambda_o\kappa_o)_+\rho^{2}\|g_o\|^{2}+\sum_{o\ \mathrm{product}}|\lambda_o|\rho^{2}\|g_{o,x}\|\|g_{o,y}\|.
\end{equation}
\end{lemma}

The quantity $s_{\mathrm{spec}}$ is the exact second-order maximum of $\Psi$, and $s_{\mathrm{op}}$ is, to
leading order in $\rho$, the right side of the lower bound~\eqref{eq:Tlower} with the adjoints and the linearized half-widths. It is the size that Proposition~\ref{prop:T} predicts for a per-operation bound. The ratio $s_{\mathrm{op}}/s_{\mathrm{spec}}$ is therefore the predicted looseness of bound propagation on a second-order part. It factors into the effective rank $\tr(H_0^{+})/\lambda_{\max}(H_0)$ and the loss from bounding each operation without cancellation. The second inequality of the bound~\eqref{eq:chain} replaces each product term by two rank-one terms, through a matrix form of the arithmetic--geometric mean inequality (Appendix~\ref{app:proofs}).

\section{Computational Results}\label{sec:experiments}

The theory makes predictions that experiments can test. The gaps should have the orders in the perturbation radius that the theorems predict, and the value-weighted secant slack, rather than the score width or the raw slack, should govern the accuracy of the certificate. Bound propagation should be loose by at least the factor that the per-operation analysis predicts. This section tests these predictions on
synthetic heads, where the regimes of the theory can be probed one at a time, and on every head
of GPT-2 small, against exact worst cases and against linear bound propagation.

All computations are in double precision. The semidefinite relaxations are solved with MOSEK~11~\cite{mosek2026mosek} and the exact references on synthetic heads with Gurobi~13~\cite{gurobi2026gurobi} as nonconvex programs (Appendix~\ref{app:relaxation}). The transformer weights are read with
PyTorch and the Hugging Face library~\cite{paszke2019pytorch,wolf2020transformers}. Interval bound
propagation (IBP), CROWN and \acrown{} are those of
auto\_LiRPA~0.8.0~\cite{xu2020automatic,zhang2018efficient,xu2021fast}, run in double precision with the default options of the library. These bound the softmax as one operation, by the convex bounds of Wei et al.~\cite{wei2023convex}.
The experiments ran on one workstation with an Intel Xeon w5-2465X processor (16 cores, 3.1 GHz) and 256 gigabytes of memory,
without graphics processors. Gaps are measured against exact worst cases where available and
otherwise against the best feasible value from multistart local ascent, whose starting points are listed in Appendix~\ref{app:relaxation}. Every semidefinite value reported for GPT-2 is the safe dual bound of Proposition~\ref{prop:safe}.

The synthetic heads have $d=8$, $d_k=d_v=4$ and $n\in\{2,3,4\}$ keys, with balls or boxes of common
radius $\rho\in[0.01,1]$ around nominal tokens, of norm $\sqrt d$ in the random and rank-one families. Five families probe the regimes of the theory. The first consists of random heads at attention temperatures $\|M\|\in\{0.1,0.3,1,3,10\}$, and the second of heads with a dominant key of margin $1$ to $8$. In the ambiguous heads of the third family, two keys tie in score but differ strongly in value along the specification. The fourth family has a rank-one query--key matrix. The fifth is a two-key family of the type of the sharpness instances of Proposition~\ref{prop:sharp}, in which one key raises its score while it lowers its value. Its values are proportional to the displacement, so that the slopes of all its gaps in the perturbation radius are larger by one. This gives $8{,}100$ instances, each certified by every method. The reference is a global solve with a time limit of $300$ seconds for $\rho\le0.2$ and $90$ seconds for larger radii, except on $40$ three-key instances of the rank-one family at $\rho=0.2$, whose reference is the best feasible value. The synthetic runs use the wider log-attention intervals $[\slo_j-\tauhi,\shi_j-\taulo]$, which contain the log-attention intervals~\eqref{eq:tint}, and report the solver's primal value, which differs from the safe value by at most $4\times10^{-8}$. A smaller rerun with the intervals~\eqref{eq:tint} gives the same slopes. The \emph{interval pattern bounds} combine interval bounds on the pattern with the exact support of each value.

\begin{figure}[t]
\centering
\includegraphics[width=\textwidth]{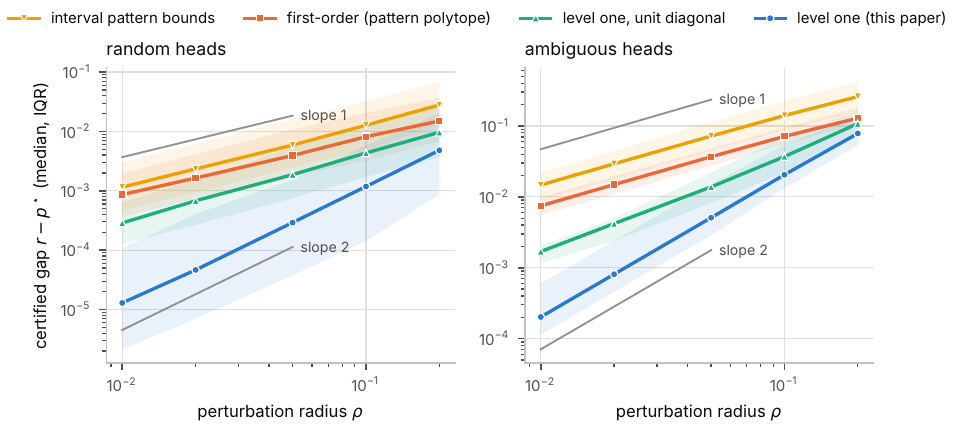}
\caption{Gap of the interval pattern bounds, the first-order certificate and the level-one
relaxation with and without its interval diagonal (R4) against the perturbation radius, on random
and ambiguous synthetic heads with ball token sets ($n=2$; medians and interquartile ranges over
$10$ seeds and all configurations).}
\label{fig:synthetic}
\end{figure}

\begin{table}[t]
\caption{Slope of the certified gap in the perturbation radius on synthetic heads with ball token
sets, $\rho\le0.1$ (medians of per-series least-squares fits, with interquartile ranges).}
\label{tab:slopes}
\begin{tabular*}{\textwidth}{@{\extracolsep\fill}lcccc}
\toprule
method & random & ambiguous & margin & rank-one\\
\midrule
interval pattern bounds & $1.02\,[1.00,1.06]$ & $1.01\,[0.98,1.02]$ & $1.04\,[0.96,1.09]$ & $1.01\,[0.98,1.06]$\\
first order, $\Rpp$ & $1.00\,[0.94,1.01]$ & $0.99\,[0.98,0.99]$ & $1.02\,[0.86,1.12]$ & $1.00\,[0.96,1.03]$\\
level one, unit diagonal & $1.10\,[1.03,1.33]$ & $1.35\,[1.24,1.54]$ & $1.28\,[1.18,1.44]$ & $1.14\,[1.03,1.41]$\\
level one, $\Rlift$ & $2.01\,[1.93,2.01]$ & $2.00\,[1.99,2.01]$ & $2.02\,[1.80,2.05]$ & $2.00\,[2.00,2.02]$\\
\botrule
\end{tabular*}
\end{table}

The orders of accuracy are confirmed first. Figure~\ref{fig:synthetic} and Table~\ref{tab:slopes}
report the slope of $\log(r-\pstar)$ against $\log\rho$ on $\rho\le0.1$. The first-order certificate
and the interval pattern bounds have slopes between $0.99$ and $1.04$ on the four families of the table, and the level-one relaxation between $2.00$ and $2.02$, on balls and on boxes alike. This is what the first- and second-order gap bounds of Theorems~\ref{thm:B} and~\ref{thm:F0v} predict. With the unit diagonal $Y[a_j,a_j]\le\bar a_j$ in
place of (R4), the slopes fall to between $1.10$ and $1.35$. The gaps are then essentially of first
order and eventually exceed any bound of second order, consistent with the role of (R4) in the
proof. At fixed radius the gap of the level-one relaxation has order $1.97$ to $2.03$ in
the temperature $\|M\|$ on random heads.

The regimes appear as the secant width grows with the temperature $\|M\|$ at fixed radius. On random heads with three keys, the level-one
certificate is $125$ times tighter than the first-order one at $w_t=0.05$, six times at $w_t=1$ and still $1.6$ times at $w_t=8$. All methods are loose in the band $1\le w_t\le8$, where several keys compete with comparable scores. On half of the seeds, one key dominates at every width. Its certified attention weight grows with $w_t$ and exceeds $0.9$ on all five of these seeds from $w_t=12$ on. On them the first-order certificate becomes essentially exact at large widths. The interval pattern bounds are $9.5$ to $28$ times looser than the level-one certificate, depending on the family and the number of keys.
Taking the minimum of the two certificates, as the certificate of Definition~\ref{def:cert} does,
costs a median $0.2$ seconds per synthetic instance.

The test on a trained model uses every head of GPT-2 small on five plain text prompts of $19$ to $22$ tokens, listed in Appendix~\ref{app:relaxation}. The query positions are $i=4,8,12$ and the last position, counted from zero so that position $i$ attends to $i+1$ keys. The token sets are balls in the post-normalization space
around the actual inputs of the head, with radius $\rho_j=\rhorel\,\|\bar z_j\|$ proportional to each
token's norm, for relative radii $\rhorel$ between $0.001$ and $0.05$. Every token in the context, including the query, is perturbed. The specification is the direction of the head's nominal output.
A head at a given prompt, position and radius is a \emph{unit}. Slopes in the radius are fitted for each head, prompt and position and are called per-unit slopes. The level-one certificate was computed on $6{,}480$ units at the positions $4$, $8$ and $12$, and the last position enters the exact computations with a fixed query.

The attention logits of GPT-2 are sensitive to perturbations of the normalized tokens, and this
places trained heads in a regime of their own. The median secant width is $240$ to $265$ times $\rhorel$ (Table~\ref{tab:gpt2}),
so the second-order regime $w_t\le1$ corresponds to perturbations of less than about $0.4\%$ of a token's norm. At $\rhorel=0.01$, $94\%$ of the units lie in the band $1\le w_t\le8$ in which all methods are loose on synthetic heads, and only two of the $6{,}480$ units have a certified dominant key.

\begin{table}[t]
\caption{Certificate accuracy on GPT-2 small by relative radius, with the relative and
range-normalized gaps of Section~\ref{sec:problem}.}
\label{tab:gpt2}
\begin{tabular*}{\textwidth}{@{\extracolsep\fill}lccccccc}
\toprule
$\rhorel$ & units & median $w_t$ & \multicolumn{2}{c}{median relative gap} & range-norm. gap & \multicolumn{2}{c}{share of units}\\
\cmidrule{4-5}\cmidrule{7-8}
 & & & certificate & first order & certificate & $\le1\%$ & $\le10\%$\\
\midrule
$0.001$ & $720$ & $0.24$ & $0.0093$ & $0.205$ & $1.8\times10^{-4}$ & $54\%$ & $100\%$\\
$0.002$ & $720$ & $0.48$ & $0.019$ & $0.213$ & $7.3\times10^{-4}$ & $32\%$ & $100\%$\\
$0.005$ & $720$ & $1.21$ & $0.048$ & $0.233$ & $4.8\times10^{-3}$ & $16\%$ & $88\%$\\
$0.01$ & $2{,}160$ & $2.65$ & $0.135$ & $0.322$ & $0.030$ & $7\%$ & $37\%$\\
$0.02$ & $2{,}160$ & $5.24$ & $0.212$ & $0.349$ & $0.097$ & $7\%$ & $24\%$\\
\botrule
\end{tabular*}
\end{table}

Table~\ref{tab:gpt2} summarizes the accuracy by radius. Radii up to $0.005$ use query position
four, and the radii $0.01$ and $0.02$ use positions four, eight and twelve. In the second-order regime,
$\rhorel\le0.002$, every unit is certified within $10\%$ of its range of motion, and at
$\rhorel=0.001$ more than half within $1\%$. In the band, the level-one certificate remains a
median factor two tighter than the first-order one and is strictly tighter on $99.5\%$ of units. At $\rhorel=0.01$ it pins the worst case to $3\%$ of the nominal spread of the head's output.

\begin{figure}[t]
\centering
\includegraphics[width=\textwidth]{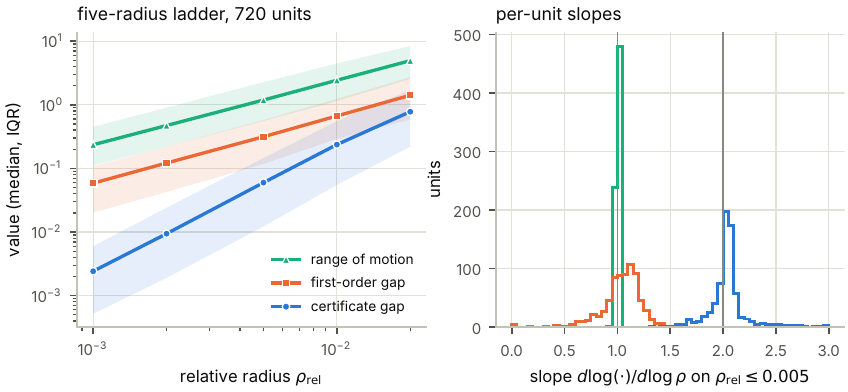}
\caption{The order of accuracy on GPT-2 small (all heads, five prompts, query position four).
Left: gaps of the two certificates and the output's range of motion against the relative radius
(medians and interquartile ranges over $720$ units). Right: per-unit slopes on $\rhorel\le0.005$.}
\label{fig:ladder}
\end{figure}

The decisive test of the orders of accuracy of Corollary~\ref{cor:orders} is the dependence of the gap on the radius within
each unit. For all $144$ heads and five prompts at position four, each head and prompt was certified at
the same five radii from $0.001$ to $0.02$ (Figure~\ref{fig:ladder}). On the three smallest radii, the median
per-unit slope of the certificate's gap is $2.04$ (interquartile range $1.99$ to $2.08$), against
$1.06$ for the first-order certificate and $1.00$ for the output's range of motion, as the theory
predicts. The slopes of the certificate are taken over the $716$ units with at least two gaps above
the numerical floor. The slopes between consecutive radii stay near two up to $\rhorel=0.01$ and fall to $1.83$
between $0.01$ and $0.02$, where the units enter the band. Measured through tightness radii at two
tolerances on three heads per layer, the exponent of the certificate is $0.52$ and that of the
first-order certificate $0.87$, against the predictions $1/2$ and $1$, respectively. A tightness
radius beyond the largest radius tested enters these medians at that radius. The measured gaps are the certificate's own. Stronger ascents,
including one that alternates the exact maximization over the keys of Theorem~\ref{thm:exactfq} with
local ascent over the query, improve the feasible value by at most $1.4\times10^{-9}$ of the spread.
Where the band leaves a gap, branching closes most of it. The relative bracket is the certificate minus the best feasible value, divided by the range of motion. On prompt zero at position four, with $24$ semidefinite solves per head and the search rules of Appendix~\ref{app:relaxation}, the median relative bracket falls from $0.087$ to $0.020$ on the $144$ heads at $\rhorel=0.01$. On the $24$ hardest heads at $\rhorel=0.002$, it falls from $0.037$ to $0.0049$. Its lower end stays fixed, so the whole decrease removes slack of the root relaxation. These runs split the key with the largest weighted secant slack $S_j$. On $72$ of the heads at $\rhorel=0.01$, the bound-lookahead rule reaches a median bracket of $0.013$ instead of $0.029$.

\begin{figure}[t]
\centering
\includegraphics[width=\textwidth]{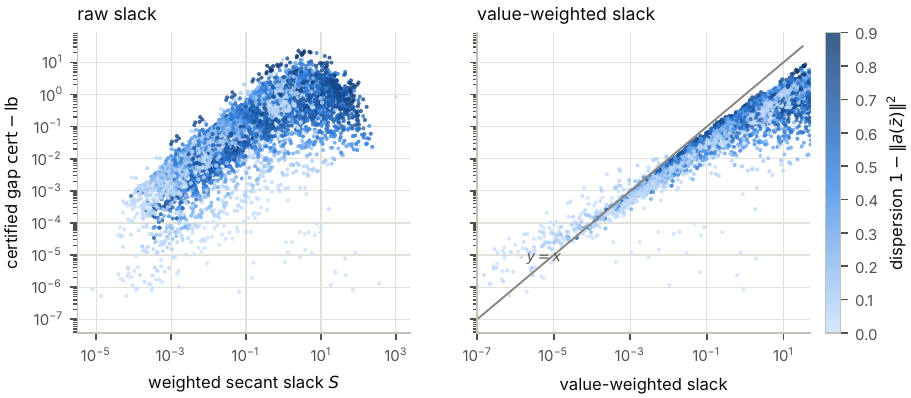}
\caption{The gap of the certificate on GPT-2 small against the weighted secant slack $S$ (left)
and against the value-weighted slack (right), coloured by the dispersion of the nominal pattern;
$6{,}442$ units.}
\label{fig:collapse}
\end{figure}

The second-order bound of Theorem~\ref{thm:F0v} predicts that the gap is governed by the value-weighted secant slack, not by the score width or the raw slack. Figure~\ref{fig:collapse} tests this on the $6{,}442$ level-one units
whose gap exceeds the numerical floor. In the raw plot, units with concentrated attention lie several
orders of magnitude below units with dispersed attention at the same slack. The right panel replaces $S$ by the value-weighted slack
\begin{equation*}
  \sum_jS_j\pos{\hmax_j-\hat\varphi_{\min}},
\end{equation*}
with $\hat\varphi_{\min}$ the smallest output found by local search, and colours each unit by the
dispersion $1-\|a(\bar z)\|^{2}$ of its nominal pattern. After value weighting, the structure
disappears on the units with $S<1$, because the slack of concentrated heads sits on keys with small
attention whose values cannot exceed the output. The gap lies below the diagonal on $97\%$ of
the units. The rank correlation of the gap with $S$ is $0.76$, and with the value-weighted slack of the
figure it is $0.91$. The knapsack term of the second-order bound~\eqref{eq:F0v} itself uses the exact
secant gaps, the slack budget and the certified lower bound $\phimin$. Its rank correlation with the
gap is $0.93$ overall, $0.99$ on the units with $S<1$ and $0.74$ on those with $S\ge1$. The gap is a median $21\%$
and at most $56\%$ of the second-order bound.

\begin{table}[t]
\caption{Exact relative gaps with a fixed query on GPT-2 small (all $144$ heads, five prompts, query
position four; $720$ units per radius).}
\label{tab:exactfq}
\begin{tabular*}{\textwidth}{@{\extracolsep\fill}lcccc}
\toprule
$\rhorel$ & \multicolumn{2}{c}{certificate} & first order & share of units\\
\cmidrule{2-3}
 & median [IQR] & max & median & within $1\%$\\
\midrule
$0.002$ & $0.0039\ [0.0012,0.0070]$ & $0.020$ & $0.172$ & $89\%$\\
$0.005$ & $0.010\ [0.003,0.018]$ & $0.056$ & $0.181$ & $50\%$\\
$0.01$ & $0.022\ [0.008,0.039]$ & $0.13$ & $0.193$ & $32\%$\\
$0.02$ & $0.048\ [0.018,0.084]$ & $0.28$ & $0.215$ & $15\%$\\
\midrule
slope & \multicolumn{2}{c}{$2.04\ [2.01,2.07]$} & $1.04$ & \\
\botrule
\end{tabular*}
\end{table}

Where it applies, namely for a head whose reading position is protected, the exact fixed query solution of Theorem~\ref{thm:exactfq} removes the remaining caveat that the reference is only a feasible value. Every head of GPT-2 small at the four query positions, five prompts and six relative radii was solved exactly with the query fixed. These are $17{,}280$ units, with certified brackets narrower than $10^{-10}$ of the nominal spread of the values along the specification, a median $6\times10^{-10}$ of the output's range of motion. The computation bounds each $\mu_u$ by a certified subdivision of the boundary angle of its ellipse, in place of the closed form of Theorem~\ref{thm:exactfq}(iii). Measured against the truth, the order of accuracy is the predicted one (Table~\ref{tab:exactfq}).
There the relative gap divides the gap against the exact worst case by the exact range of motion,
and the slopes are per-unit slopes of the logarithm of the gap against that of $\rhorel$ between
$0.002$ and $0.005$. On the $2{,}880$ units at position four, the relative gap of the certificate is $0.004$ at $\rhorel=0.002$ and $0.048$ at $0.02$. Its per-unit slope is $2.04$, against $1.04$ for the first-order certificate and $1.00$ for the exact range of motion. The gap is at most $0.78$ of the second-order bound~\eqref{eq:F0v}. Multistart ascent finds the exact worst case to $10^{-9}$ of the nominal spread of the values on every one of these units, so with a fixed query the whole gap is the relaxation's. Letting the reading position move more than doubles the swing of the logits, since the secant width with a varying query
is a median $2.36$ to $2.42$ times that with a fixed query, depending on the radius.

\begin{table}[t]
\caption{Relative gaps of linear bound propagation and of our certificates on the same GPT-2 units
(all $144$ heads, five prompts, query position four; $720$ units per row; medians). The last two
columns are median per-unit ratios to the gap of the certificate.}
\label{tab:crown}
\footnotesize
\begin{tabular*}{\textwidth}{@{\extracolsep\fill}llccccccc}
\toprule
query & $\rhorel$ & IBP & CROWN & \acrown{} & first & certifi- & \multicolumn{2}{c}{ratio to cert.}\\
\cmidrule{8-9}
 & & & & & order & cate & CROWN & $\alpha$\\
\midrule
fixed & $0.002$ & $5.7$ & $0.020$ & $0.016$ & $0.172$ & $0.0039$ & $4.4$ & $3.9$\\
      & $0.005$ & $7.4$ & $0.053$ & $0.042$ & $0.181$ & $0.010$ & $4.5$ & $3.9$\\
      & $0.01$ & $11.2$ & $0.113$ & $0.090$ & $0.193$ & $0.022$ & $4.5$ & $3.8$\\
      & $0.02$ & $16.8$ & $0.255$ & $0.199$ & $0.215$ & $0.048$ & $4.7$ & $4.0$\\
varying & $0.001$ & $7.2$ & $0.083$ & $0.072$ & $0.205$ & $0.0093$ & $7.8$ & $6.7$\\
      & $0.002$ & $9.1$ & $0.186$ & $0.159$ & $0.213$ & $0.019$ & $8.6$ & $7.5$\\
      & $0.005$ & $15.3$ & $0.636$ & $0.527$ & $0.233$ & $0.048$ & $11.8$ & $9.8$\\
      & $0.01$ & $19.2$ & $2.17$ & $1.72$ & $0.269$ & $0.098$ & $22.2$ & $17.5$\\
\botrule
\end{tabular*}
\end{table}

\begin{figure}[t]
\centering
\includegraphics[width=\textwidth]{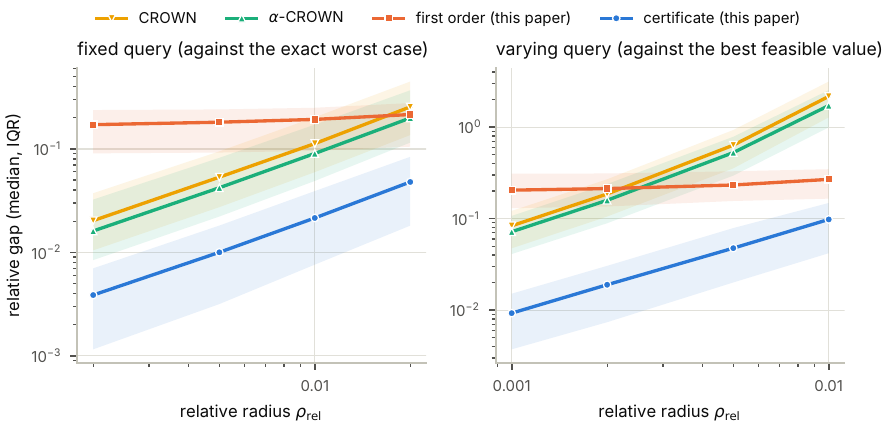}
\caption{Relative gaps of CROWN, \acrown{}, the first-order certificate and the certificate on the same
GPT-2 units ($720$ units per radius; medians and interquartile ranges).}
\label{fig:crown}
\end{figure}

Linear bound propagation comes with no order of accuracy, and the comparison with it is made on exactly the same footing. It uses the same heads, sets and specifications, with $2{,}880$ fixed query units measured against the exact worst case and $2{,}880$ varying query units measured against the best feasible value. The head is given to the bound propagation library in the same reduced coordinates, with each perturbed token as a separate input in its own ball. Table~\ref{tab:crown} and Figure~\ref{fig:crown} give the result.
Interval bound propagation is vacuous, with relative gaps between $5.7$ and $19$, and is omitted
from Figure~\ref{fig:crown}. CROWN and \acrown{} are empirically of second order, with a
per-unit slope of $2.11$ with a fixed query, the same as the certificate, and $2.39$ with a varying one, against $2.05$ for the certificate. These slopes are fitted over the four radii of each query type, and over $0.002$--$0.005$ alone the certificate's slope is the $2.04$ of the exact comparison in Table~\ref{tab:exactfq}. The constant of bound propagation is larger. With a fixed query their gap is $3.8$
to $4.7$ times the certificate's at every radius, and with a varying query the ratio grows from $7.8$ ($6.7$ for \acrown{}) at
$\rhorel=0.001$ to $22$ ($17.5$) at $0.01$. The certificate is tighter on every one of the $5{,}760$ units, by at
least a factor two on $96\%$ of them. Bound propagation is fast, $0.01$ to $0.02$ seconds per unit for CROWN and $0.2$ to $0.7$ seconds for \acrown{}, against a median of $10$ seconds for the semidefinite certificate with a fixed query and $22$ seconds with a varying one. At the smallest radii it is tighter than the first-order certificate, so it is the better fast bound there.
More optimization does not change the picture. On the $1{,}152$ units of one prompt, $20$ to $500$ iterations of \acrown{} at two learning rates move the median ratio of its gap to the certificate's by less than $0.1\%$. This is what the per-operation lower bound of Proposition~\ref{prop:T} leads one to expect.

The last test asks whether the looseness of bound propagation is the sum of per-operation maxima that Proposition~\ref{prop:T} predicts, on a function where both sides can be computed. For a block $\ell$ of GPT-2 and a head of block $\ell+1$, let one token at the input of block $\ell$ be perturbed in a ball of relative radius $\rhorel$. Let $F$ map the residual stream at the input of block $\ell$ to the normalized tokens read by the head, and let $f$ be the remainder of $F$ read along the direction to which the head is sensitive. Namely, $f$ is the inner product of the remainders of the head's queries, keys and values with the gradient, with respect to them, of the head's nominal output along the specification. The function $f$ vanishes with its gradient at the nominal input, and its maximum is, to first order in the head, the amount by which the second-order part of the block can move the head's output. On $132$ units (blocks $0$--$1$, $3$--$4$, $6$--$7$ and $9$--$10$, the $12$ downstream heads, $\rhorel\in\{0.0005,0.001,0.002\}$, without blocks $0$--$1$ at the largest radius), CROWN was run on the network that computes $f$, and the attained maximum of $f$ was found by multistart ascent. The Hessian
$H_0$ of $f$ was computed with its decomposition~\eqref{eq:H0} over the unary operations and products of the block, with the softmax written through exponentials, a sum, a reciprocal and products.

\begin{figure}[t]
\centering
\includegraphics[width=0.6\textwidth]{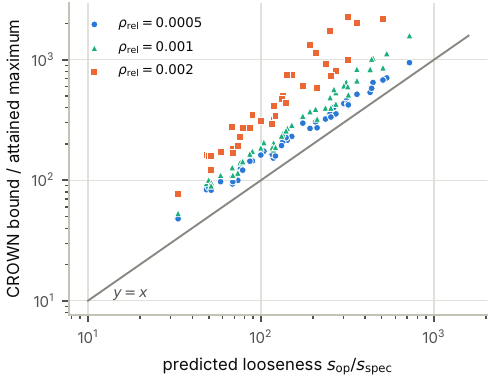}
\caption{Looseness of CROWN on the remainder of a GPT-2 block, its bound on $\max f$ divided by the
attained maximum, against the looseness $s_{\mathrm{op}}/s_{\mathrm{spec}}$ predicted by
Lemma~\ref{lem:chain} ($132$ units).}
\label{fig:mechanism}
\end{figure}

Three predictions, stated before the computation, were confirmed at $\rhorel\le0.001$. The attained
maximum of $f$ equals the exact second-order term $s_{\mathrm{spec}}$ to within $0.04\%$ to $1.5\%$, so the remainder is quadratic at these radii. CROWN's bound is $1.26$ to $2.34$ times the per-operation sum $s_{\mathrm{op}}$
(median $1.63$). It is $1.11$ to $1.54$ times the per-operation terms of the lower
bound~\eqref{eq:Tlower} evaluated on CROWN's own operations and intervals, which after the attention
exceed their linearized widths. And the rank correlation of CROWN's looseness with the predicted
looseness $s_{\mathrm{op}}/s_{\mathrm{spec}}$ is $0.98$ on the $96$ units with $\rhorel\le0.001$
(Figure~\ref{fig:mechanism}), and $0.90$ over all $132$ units. CROWN bounds
the softmax as one operation, whereas $s_{\mathrm{op}}$ writes it through exponentials, a sum, a
reciprocal and products. Counting the softmax as one operation multiplies $s_{\mathrm{op}}$ by a factor
between $0.75$ and $1.01$ and leaves the rank correlation at $0.98$. Over all $132$ units the predicted looseness is large, a median $140$ (between $33$ and $722$), and its sources can be named. The effective rank $\tr(H_0^+)/\lambda_{\max}(H_0)$ is a median $11$ (between $2.4$ and $50$). The per-operation sum exceeds the trace by a further factor of about $14$,
because terms of opposite sign do not cancel and the two operands of a product are not aligned. With the softmax counted as one operation, the
operations of the tanh-approximated activation of the multilayer perceptron and the products of
attention weights and values carry $79\%$ to $92\%$ of the per-operation sum. At the radii where the remainder is quadratic, bound propagation is thus loose on these computations by at least the per-operation excess that the analysis predicts, and by at most $2.34$ times the excess of Lemma~\ref{lem:chain}. Its single-head looseness of Table~\ref{tab:crown} is consistent with the same phenomenon at a smaller scale.

\section{Conclusion}\label{sec:conclusion}

Attention is certifiable to second order, and the theory of this paper says when. The difficulty
lies not in the softmax but in the bilinear interaction of a varying query with the keys, which
over boxes makes certification NP-hard and, once the attention logits can move by a constant,
inapproximable. A fixed query or listed substitutions reduce the worst case to a separable
equation. Elsewhere, a greedy linear program certifies to first order, and an exponential cone
moment relaxation certifies to second order, an order that cannot be improved, with a gap bounded
through the value-weighted secant slack. Relaxations that treat one operation at a time pay a sum
of second-order terms instead of one coupled term. On GPT-2 small, the certificate attains the
predicted order against exact worst cases with the query fixed and is tighter than linear bound
propagation on every head. The looseness of bound propagation on block remainders tracks the
predicted excess.

The analysis concerns one head, with continuous perturbations of the tokens it reads and a
specification read linearly from its output, and three directions extend it. The first is
certification through depth, with bounds that couple the second-order terms of a block's
operations as the moment relaxation does within a head. The per-operation excess measured on
GPT-2 blocks, two orders of magnitude, marks what such bounds could recover. The second is scale.
A sparser relaxation keeps the second-order bound while its size grows only linearly with the
number of keys. The third turns the analysis into design. The
constant that controls the second-order accuracy of a head is explicit and differentiable almost
everywhere, so certifiability at a prescribed radius can be imposed during training. The accuracy
that a model must give up to be certified then becomes a well-posed question.

\bmhead{Code and data availability}
The implementation of the certificates is available at
\url{https://github.com/chimera1001/conic-attention-cert}. The repository also contains the exact
worst cases with a fixed query computed for all heads of GPT-2 small.

\clearpage
\bibliography{refs}

\clearpage
\begin{appendices}
\section{Proofs}\label{app:proofs}

This appendix proves the results of the body in the order in which they appear. Throughout,
for a probability vector $a$ and a vector $v$ we write $\bar v_a=a^\top v$, and $\Lambda$ denotes
the logistic function, namely
\begin{equation*}
  \Lambda(x)=\frac{1}{1+e^{-x}} .
\end{equation*}

\begin{proof}[Proof of Lemma~\ref{lem:lip}]
Let $a=\sigma(s)$ and let $G(s)=\diag(a)-aa^\top$ be the Jacobian of the softmax. For
$u\in\R^{n}$ the components of $G(s)u$ are $a_k(u_k-\bar u_a)$, so
\begin{equation}\label{eq:mad}
  \|G(s)u\|_1=\sum_ka_k\,|u_k-\bar u_a|,
\end{equation}
the mean absolute deviation of $u$ under $a$. Put
\begin{equation*}
  \alpha=\max_ku_k-\bar u_a,
  \qquad
  \beta=\bar u_a-\min_ku_k,
  \qquad
  p=\sum_{k:\,u_k>\bar u_a}a_k .
\end{equation*}
Since the deviations have mean zero,
\begin{equation}\label{eq:madsplit}
  \sum_ka_k|u_k-\bar u_a|=2\sum_{u_k>\bar u_a}a_k(u_k-\bar u_a)=2\sum_{u_k\le\bar u_a}a_k(\bar u_a-u_k)
  \le2\min\bigl\{p\alpha,(1-p)\beta\bigr\}.
\end{equation}
The minimum of $p\alpha$ and $(1-p)\beta$ is largest when they are equal, so
\begin{equation}\label{eq:madbound}
  \|G(s)u\|_1\le\frac{2\alpha\beta}{\alpha+\beta}\le\frac{\alpha+\beta}{2}=\tfrac12\range(u),
\end{equation}
where the second inequality is $4\alpha\beta\le(\alpha+\beta)^{2}$. Integrating along the
segment from $s'$ to $s$,
\begin{equation}\label{eq:lipint}
  \sigma(s)-\sigma(s')=\int_0^1G\bigl(s'+\theta(s-s')\bigr)(s-s')\,d\theta ,
\end{equation}
and applying the bound~\eqref{eq:madbound} with $u=s-s'$ for every $\theta$ gives the first inequality of the Lipschitz bound~\eqref{eq:lip}, namely
$\|\sigma(s)-\sigma(s')\|_1\le\tfrac12\range(s-s')$. The second holds because $\range(v)\le2\|v\|_\infty$.
The constant cannot be lowered. For $n=2$, $s=(x,0)$ and $s'=(0,0)$,
\begin{equation*}
  \|\sigma(s)-\sigma(s')\|_1=2\bigl|\Lambda(x)-\tfrac12\bigr|=\tfrac x2+O(x^{3}),
  \qquad
  \range(s-s')=x ,
\end{equation*}
so the ratio of $\|\sigma(s)-\sigma(s')\|_1$ to $\range(s-s')$ tends to $\tfrac12$ as $x\to0$, although it never equals $\tfrac12$.
\end{proof}

\begin{proof}[Proof of Lemma~\ref{lem:intervals}]
The three parts are proved in turn.
\begin{enumerate}[label=\normalfont(\roman*), leftmargin=*]
\item Each weight can be written as
\begin{equation*}
  \sigma_j(s)=\Bigl(1+\sum_{l\ne j}e^{s_l-s_j}\Bigr)^{-1},
\end{equation*}
which is increasing in $s_j$ and decreasing in every $s_l$, $l\ne j$. Its extreme values
over the box are therefore attained at the two corners that appear in the attention
intervals~\eqref{eq:aint}. The largest value $\ahi_j$ has $s_j$ at its upper end and the other
scores at their lower ends, and the smallest value $\alo_j$ has the reverse.
\item For $l\ne j$,
\begin{equation*}
  \shi_l-\slo_j=(s^{\mathrm{mid}}_l-s^{\mathrm{mid}}_j)+\tfrac12(\Delta_l+\Delta_j)\le(s^{\mathrm{mid}}_l-s^{\mathrm{mid}}_j)+\Delta .
\end{equation*}
With $q=\sum_{l\ne j}e^{s^{\mathrm{mid}}_l-s^{\mathrm{mid}}_j}$, so that $\hat a_j=(1+q)^{-1}$,
\begin{equation}\label{eq:alobound}
  \alo_j\ge\frac{1}{1+e^{\Delta}q}\ge\frac{e^{-\Delta}}{1+q}=e^{-\Delta}\hat a_j ,
\end{equation}
because $1+e^{\Delta}q\le e^{\Delta}(1+q)$. The bound $\ahi_j\le e^{\Delta}\hat a_j$ follows in
the same way from $\slo_l-\shi_j\ge(s^{\mathrm{mid}}_l-s^{\mathrm{mid}}_j)-\Delta$. Summing over the keys,
\begin{equation*}
  \sum_j\ahi_j\le e^{\Delta},
  \qquad
  \sum_j(\ahi_j-\alo_j)\le(e^{\Delta}-e^{-\Delta})\sum_j\hat a_j=2\sinh\Delta .
\end{equation*}
\item Let $x=\sum_{j\ne j^\star}e^{s_j-s_{j^\star}}$, so that $x\le(n-1)e^{-\gamma}$. Then,
\begin{equation*}
  \|\sigma(s)-e_{j^\star}\|_1=2\bigl(1-\sigma_{j^\star}(s)\bigr)=\frac{2x}{1+x},
\end{equation*}
which is increasing in $x$. With $x\le(n-1)e^{-\gamma}$ this gives the margin
bound~\eqref{eq:margin}, namely
\begin{equation*}
  \|\sigma(s)-e_{j^\star}\|_1\le\frac{2(n-1)e^{-\gamma}}{1+(n-1)e^{-\gamma}} .
\end{equation*}
\end{enumerate}
\end{proof}

\begin{proof}[Proof of Lemma~\ref{lem:secant}]
The function $\ell(t)-e^{t}$ is concave and vanishes at both endpoints, so it is nonnegative
on the interval and its maximum is attained where its derivative $m-e^{t}$, with
$m=(e^{\thi}-e^{\tlo})/w$ the slope of $\ell$, vanishes, at $t^\star=\log m$. By the mean value
theorem, $m=e^{\xi}$ for some $\xi\in(\tlo,\thi)$, so
$t^\star$ lies in the interval. Writing $m=e^{\tlo}(1+y)$ with $1+y=(e^{w}-1)/w$,
\begin{equation}\label{eq:Dderive}
  D=\ell(t^\star)-e^{t^\star}=e^{\tlo}+m\,(t^\star-\tlo)-m=e^{\tlo}\bigl(1+(1+y)\log(1+y)-(1+y)\bigr),
\end{equation}
which is the closed form $D=e^{\tlo}\bigl((1+y)\log(1+y)-y\bigr)$ of the exact secant
gap~\eqref{eq:Dclosed}. The bound by $\frac{w^{2}}{8}e^{\thi}$ is the
classical interpolation error of a convex function whose second derivative is at most
$e^{\thi}$ on the interval, and the bound by $e^{\thi}-e^{\tlo}$ holds because $\ell\le e^{\thi}$
and $e^{t}\ge e^{\tlo}$ there. Finally,
\begin{equation*}
  y=\tfrac w2+O(w^{2}),
  \qquad
  (1+y)\log(1+y)-y=\tfrac{y^{2}}{2}+O(y^{3}),
\end{equation*}
which gives the expansion~\eqref{eq:Dasym}, namely $D=\frac{w^{2}}{8}e^{\tlo}(1+O(w))$.
\end{proof}

\begin{proof}[Proof of Proposition~\ref{prop:P7v}]
Put $q_j=e^{t_j}$ and $Z=\sum_jq_j$. Since $a_j=q_j+d_j$ and $\sum_ja_j=1$,
\begin{equation*}
    \sum_jd_j=1-Z,
  \qquad
  \sigma_j=\frac{q_j}{Z},
  \qquad
  \sum_jq_jh_j=Z\,\bar h_\sigma ,
\end{equation*}
and therefore
\begin{equation}\label{eq:P7vproof}
  \sum_j(a_j-\sigma_j)h_j=Z\,\bar h_\sigma+\sum_jd_jh_j-\bar h_\sigma
  =\sum_jd_jh_j-(1-Z)\,\bar h_\sigma=\sum_jd_j\bigl(h_j-\bar h_\sigma\bigr).
\end{equation}
Under the secant bounds, the exact secant gap of Lemma~\ref{lem:secant} gives
\begin{equation*}
  0\le d_j=a_j-e^{t_j}\le\ell_j(t_j)-e^{t_j}\le D_j,
  \qquad
  \sum_jd_j=1-\sum_je^{t_j}\le1-\sum_je^{\tlo_j}=D_{\mathrm{tot}} .
\end{equation*}
Since $d\ge0$, the right side of the identity~\eqref{eq:P7vproof} is at most
$\sum_jd_j\pos{h_j-\bar h_\sigma}$, and $d$ satisfies the constraints $0\le d_j\le D_j$ and
$\sum_jd_j\le D_{\mathrm{tot}}$ of the knapsack~\eqref{eq:KN}, here with the weights
$\pos{h_j-\bar h_\sigma}$.
\end{proof}

The reductions for the complexity results use the infinity-to-one norm of a matrix $Q\in\R^{d_1\times d_2}$, namely
\begin{equation}\label{eq:inf1}
  \Gamma^\star(Q)=\|Q\|_{\infty\to1}=\max\bigl\{u^\top Qu':\ u\in[-1,1]^{d_1},\ u'\in[-1,1]^{d_2}\bigr\},
\end{equation}
where the maximum over the boxes equals the maximum over their vertices because the form is
affine in each argument. Alon and Naor construct in polynomial time, from a graph $G$ with edge
set $E$, a matrix $A$ with $2|E|$ rows and entries in $\{-1,0,1\}$
satisfying~\cite[Proposition~3.2]{alon2006approximating}
\begin{equation*}
  \|A\|_{\infty\to1}=4\,\mathrm{MAXCUT}(G).
\end{equation*}
Two facts follow. First, deciding whether $\Gamma^\star(Q)\ge k$ for an integer matrix $Q$ and an
integer $k$ is NP-hard, because deciding whether $\mathrm{MAXCUT}(G)\ge k$ is. Second, H\aa stad proved that the maximum cut cannot be approximated in polynomial time within a ratio exceeding $16/17$ unless $\mathrm{P}=\mathrm{NP}$~\cite{hastad2001some}, and the same threshold holds for unweighted graphs~\cite{crescenzi2001weighted}, to which the construction applies. Hence, no polynomial-time algorithm
returns, for every integer matrix $Q$, a number $\hat\Gamma$ with
\begin{equation*}
  \Gamma^\star\le\hat\Gamma\le\varrho \Gamma^\star
  \qquad\text{for a fixed }\varrho<17/16,
\end{equation*}
unless $\mathrm{P}=\mathrm{NP}$, since $\hat\Gamma/(4\varrho)$ would approximate $\mathrm{MAXCUT}(G)$
within the ratio $1/\varrho>16/17$. For a nonzero integer matrix we also have
\begin{equation*}
  1\le \Gamma^\star(Q)\le T(Q)\eqdef\sum_{k,l}|Q_{kl}| .
\end{equation*}

\begin{proof}[Proof of Theorem~\ref{thm:hard}]
Let $Q\in\mathbb Z^{d_1\times d_2}$ be nonzero and let $k$ be an integer with $1\le k\le T=T(Q)$;
for other $k$ the question $\Gamma^\star(Q)\ge k$ is trivial. Let $d=d_1+d_2+2$ and split the
coordinates into a first block of $d_1$ coordinates, a second block of $d_2$ coordinates and two
flags $f_1,f_2$. The first token and the second token, which is the query, are, respectively,
\begin{equation*}
  z_1=(0,u',0,1),\quad u'\in[-1,1]^{d_2},
  \qquad
  z_2=(u,0,1,0),\quad u\in[-1,1]^{d_1},
\end{equation*}
so both vary in boxes. Let $M$ be the matrix with
\begin{equation*}
  z^\top Mz'=T^{-1}\,(\text{first block of }z)^\top Q\,(\text{second block of }z'),
\end{equation*}
let $N$ map $z$ to its $f_2$-coordinate times a fixed unit vector $e$, and let $c=e$. Then,
\begin{equation}\label{eq:hardscores}
  s_1=\frac{u^\top Qu'}{T},\qquad s_2=\frac{u^\top Q\,0}{T}=0,\qquad h_1=c^\top Nz_1=1,\qquad h_2=c^\top Nz_2=0,
\end{equation}
so the values do not depend on the varying coordinates, and
\begin{equation}\label{eq:hardphi}
  \varphi=\frac{e^{s_1}}{e^{s_1}+e^{s_2}}=\Lambda\Bigl(\frac{u^\top Qu'}{T}\Bigr).
\end{equation}
Because $\Lambda$ is strictly increasing, $\pstar=\Lambda(\Gamma^\star(Q)/T)$. Since $\Gamma^\star(Q)$ is an
integer, $\Gamma^\star(Q)\ge k$ holds if and only if $\pstar\ge\beta$ for any $\beta$ in the interval
\begin{equation*}
  \Bigl(\Lambda\bigl(\tfrac{k-1}{T}\bigr),\ \Lambda\bigl(\tfrac kT\bigr)\Bigr].
\end{equation*}
This interval lies in $\Lambda([0,1])$, where $\Lambda'\ge\Lambda'(1)>0.19$, so its length is at
least $0.19/T$, and a rational $\beta$ in it with $O(\log T)$ bits is computed in polynomial time.
The instance and $\beta$ have a size polynomial in that of $(Q,k)$, which proves the theorem.
\end{proof}

\begin{proof}[Proof of Theorem~\ref{thm:inapprox}]
Let $Q$ be a nonzero integer matrix and let $\mathrm{SDP}(Q)$ be the value of the semidefinite
relaxation of the norm~\eqref{eq:inf1} in which $u_k$ and $u'_l$ are replaced by unit vectors.
By Grothendieck's inequality with Krivine's bound on its constant~\cite{krivine1979constantes},
\begin{equation*}
  \Gamma^\star\le\mathrm{SDP}(Q)\le K_G\Gamma^\star,
\end{equation*}
where $K_G<1.783$.
Since $\mathrm{SDP}(Q)\ge \Gamma^\star\ge1$, a number $U$ with $\mathrm{SDP}(Q)\le U\le1.001\,\mathrm{SDP}(Q)$
is computed in polynomial time, and $\Gamma^\star\le U\le1.785\,\Gamma^\star$. In the two-token instance of the proof of Theorem~\ref{thm:hard}, use $U$ in place of $T$. The score of the first token is then
$\vartheta=u^\top Qu'/U$, which ranges over $[-\Gamma^\star/U,\Gamma^\star/U]$, an interval of length
between $2/1.785>1.12$ and $2$, so the instance is of the type considered in the theorem. Its worst case and the range of its output are, respectively,
\begin{equation}\label{eq:inapproxrange}
\begin{aligned}
  \pstar&=\Lambda\bigl(\Gamma^\star/U\bigr)\le\Lambda(1),\\
  \max\varphi-\min\varphi&=2\Lambda\bigl(\Gamma^\star/U\bigr)-1=\tanh\frac{\Gamma^\star}{2U}\le\tanh\tfrac12<0.463 .
\end{aligned}
\end{equation}
Suppose that an algorithm returns $r$ with $\pstar\le r\le\pstar+\delta$, where
\begin{equation*}
  \delta=\tfrac1{200}(\max\varphi-\min\varphi)<0.003 .
\end{equation*}
Then,
\begin{equation*}
  \tfrac12<r<\Lambda(1)+0.003<\Lambda(1.5),
\end{equation*}
so $\hat x=\Lambda^{-1}(r)$ lies in
$[\Gamma^\star/U,1.5)$, and since $\Lambda'\ge\Lambda'(1.5)>0.149$ on $[0,1.5]$,
\begin{equation}\label{eq:inapproxinv}
  0\le\hat x-\frac{\Gamma^\star}{U}\le\frac{\delta}{0.149}<0.021 .
\end{equation}
The number $\hat\Gamma=U\hat x$, computed in polynomial time, therefore satisfies
\begin{equation*}
  \Gamma^\star\le\hat\Gamma\le \Gamma^\star+0.021\,U\le1.038\,\Gamma^\star,
\end{equation*}
an approximation of $\|Q\|_{\infty\to1}$ within a factor $1.038<17/16$. By the second fact stated before the proof of Theorem~\ref{thm:hard}, the inapproximability of the maximum cut, this is impossible unless
$\mathrm{P}=\mathrm{NP}$.
\end{proof}

\begin{proof}[Proof of Theorem~\ref{thm:exactfq}]
The three parts are proved in turn.
\begin{enumerate}[label=\normalfont(\roman*), leftmargin=*]
\item With the query fixed, the score $s_u$ and the value $h_u$ of each varying key are affine
functions of its own token, and the pairs $(s_u,h_u)$ range over the sets $C_u$ independently of each other. Each $C_u$ is compact, as a continuous image of $\Z_u$. Every function
$\lambda\mapsto e^{s}(h-\lambda)$ is affine with slope $-e^{s}$, so $\mu_u$, a maximum of such
functions over a compact set, is convex, continuous and strictly decreasing with slopes at most $-\min_{C_u}e^{s}<0$. The fixed terms are affine and strictly decreasing, and $K_{\mathrm f}$ is
nonempty because it contains the query position. Hence, $H$ is convex, continuous and strictly
decreasing, and it tends to $+\infty$ and $-\infty$ at $-\infty$ and $+\infty$, respectively, so it has a unique
root $\lambda^\star$. For every admissible input $z$ and every $\lambda$, separability gives
\begin{equation}\label{eq:sep}
  \sum_je^{s_j(z)}\bigl(h_j(z)-\lambda\bigr)\le\sum_{k\in K_{\mathrm f}}e^{s_k}(h_k-\lambda)+\sum_{u\in K_{\mathrm v}}\mu_u(\lambda)=H(\lambda),
\end{equation}
with equality for tokens realizing maximizers of the $\mu_u(\lambda)$. Since the denominator
$\sum_je^{s_j(z)}$ is positive, $\varphi(z)\le\lambda$ whenever the left side of
inequality~\eqref{eq:sep} is nonpositive. At $\lambda=\lambda^\star$ the right side is zero, so
$\varphi(z)\le\lambda^\star$ for every input, and the maximizing tokens give equality. Therefore,
$\pstar=\lambda^\star$, and it is attained.
\item The set $C_u$ is the image of a polytope under an affine map into the plane, hence a
polygon. For a box or zonotope $\bar z+B[-\omega,\omega]$ with $r_u$ generators it is a zonogon
generated by $r_u$ segments, with at most $2r_u$ vertices obtained by sorting the generators by angle. For the hull of $m_u$ points it is the planar convex hull of their images, with at most $m_u$ vertices. From the projected generators or points, the constructions cost $O(r_u\log r_u)$ and $O(m_u\log m_u)$ operations, respectively. Let
$f(s,h)=e^{s}(h-\lambda)$. If $\max_{C_u}h>\lambda$, the maximum of $f$ is positive and attained in
$\{h>\lambda\}$, where
\begin{equation*}
  \log f(s,h)=s+\log(h-\lambda)
\end{equation*}
is concave with a gradient whose first component equals one. It has no stationary point, so the
maximum lies on the boundary of $C_u$. On an edge parametrized as
$(s_0+\alpha\tau,h_0+\beta\tau)$, $\tau\in[0,1]$, the concave function
$\alpha\tau+\log(h_0+\beta\tau-\lambda)$ is maximized at an endpoint or at the stationary point
\begin{equation*}
  h_0+\beta\tau-\lambda=-\beta/\alpha .
\end{equation*}
If $\max_{C_u}h\le\lambda$, then $f\le0$, and maximizing $f$ means minimizing $e^{s}(\lambda-h)$,
whose logarithm $s+\log(\lambda-h)$ is concave where $h<\lambda$. The minimum of a concave function
over a polygon is attained at a vertex, and the value zero is attained if $h=\lambda$ somewhere. Each
$\mu_u(\lambda)$ is therefore computed by one pass over the vertices and edges of $C_u$, and each evaluation of $H$ costs $O(|K_{\mathrm f}|+\sum_uV_u)$ operations. Since
$\pstar\in[\min_j\hmin_j,\max_j\hmax_j]$ and $H$ is strictly decreasing, bisection on $\lambda$ with
the sign of $H$ computes $\pstar$ to accuracy $\varepsilon$ in $O(\log(R/\varepsilon))$ evaluations.
\item The same case analysis applies to a filled ellipse $C_u$. If $\max_{C_u}h>\lambda$, the
maximum lies on the boundary because $\log f$ has no stationary point. If $\max_{C_u}h=\lambda$, then
$\mu_u(\lambda)=0$, attained where $h$ is largest, which includes a boundary point. Otherwise,
$h<\lambda$ on $C_u$, and $\mu_u(\lambda)$ is minus the minimum of $e^{s}(\lambda-h)$, whose
logarithm is concave. A concave function attains its minimum over a compact convex set at an
extreme point, which here lies on the boundary. Write the boundary as
$(s(\theta),h(\theta))=(s_0,h_0)+A(\cos\theta,\sin\theta)$ with $A\in\R^{2\times2}$, which covers
degenerate ellipses as well. Then,
\begin{equation*}
  \frac{d}{d\theta}f\bigl(s(\theta),h(\theta)\bigr)=e^{s(\theta)}\Bigl(s'(\theta)\bigl(h(\theta)-\lambda\bigr)+h'(\theta)\Bigr),
\end{equation*}
and the bracket is a trigonometric polynomial of degree at most two. Unless it vanishes
identically, in which case $f$ is constant on the boundary, it has at most four zeros in a period.
The substitution $x=\tan(\theta/2)$ turns them into the real roots of a polynomial of degree at most
four, and $\theta=\pi$ is checked separately. Hence, $\mu_u(\lambda)$ is the largest value of $f$ at
no more than five angles, computed in a constant number of operations, and the costs of evaluating
$H$ and of computing $\pstar$ follow as in part~(ii).
\end{enumerate}
\end{proof}

\begin{proof}[Proof of Proposition~\ref{prop:greedy}]
For $a$ in the box, every coordinate satisfies
\begin{equation*}
  |a_j-\hat a_j|=|a_j-b_j|+|b_j-\hat a_j|,
\end{equation*}
because $b_j$ is the point of $[\alo_j,\ahi_j]$ closest to $\hat a_j$; on the box the cut
therefore reads $\|a-b\|_1\le\epsilon'$. The polytope $\mathcal A$ is nonempty, since it contains
the pattern of every admissible input. Write $a-b=p-q$ with $p,q\ge0$, let $\delta=1-\sum_jb_j$
and suppose $\delta\ge0$, the other case being symmetric. Then,
\begin{equation*}
  \sum_jp_j=\delta+m,
  \qquad
  \sum_jq_j=m,
  \qquad
  \|a-b\|_1=\delta+2m
\end{equation*}
for some $m\ge0$ when $p$ and $q$ have disjoint supports. For a fixed $m$, the best $p$ fills the
keys in decreasing order of $\hmax_j$ up to their capacities $\ahi_j-b_j$ with a total mass
$\delta+m$, and the best $q$ empties the keys in increasing order of $\hmax_j$ down to their floors
$b_j-\alo_j$ with a total mass $m$. These are two fractional knapsacks, with values $F^{+}(\delta+m)$ and $F^{-}(m)$, $F^{+}$ concave and $F^{-}$ convex. If the two solutions overlap on a key, the net
transfer $p-q$ is still feasible, with the same objective and a smaller $\ell_1$ cost, so
\begin{equation}\label{eq:greedyval}
  \Rpp=\sum_j\hmax_jb_j+\max\Bigl\{F^{+}(\delta+m)-F^{-}(m):\ m\ge0,\ \delta+2m\le\epsilon'\Bigr\}.
\end{equation}
The function maximized in the expression~\eqref{eq:greedyval} for $\Rpp$ is concave and piecewise linear, with
right derivative equal to the difference of $\hmax$ between the next receiver and the next
donor, which is nonincreasing in $m$. It is maximized by increasing $m$ from zero until
$\delta+2m=\epsilon'$ or the derivative becomes nonpositive, which is the two-phase rule. Sorting
costs $O(n\log n)$ operations and each phase $O(n)$.
\end{proof}

\begin{proof}[Proof of Theorem~\ref{thm:B}]
Validity was shown after the definition of the pattern polytope certificate, Definition~\ref{def:pp}. Let $a^\star$ be optimal in the linear program~\eqref{eq:rpp}, which maximizes
$\sum_ja_j\hmax_j$ over the pattern polytope $\mathcal A$, and choose $\tilde z_j\in\Z_j$ with $h_j(\tilde z_j)=\hmax_j$ for every
key, the query being one of them; $\tilde z$ is an admissible input. Its pattern $\tilde a=\sigma(s(\tilde z))$ satisfies $\|\tilde a-\hat a\|_1\le\epsilon$ because $(\hat a,\epsilon)$ is certified, and
\begin{equation*}
  \varphi(\tilde z)=\sum_j\tilde a_j\hmax_j\le\pstar .
\end{equation*}
The midpoint $m=\tfrac12(\max_j\hmax_j+\min_j\hmin_j)$ of the range of the values satisfies $|\hmax_j-m|\le\radc$ for every $j$. Since $\sum_j(a^\star_j-\tilde a_j)=0$,
\begin{equation}\label{eq:Bproof}
\begin{aligned}
  \Rpp-\pstar&\le\sum_j\bigl(a^\star_j-\tilde a_j\bigr)\bigl(\hmax_j-m\bigr)
  \le\|a^\star-\tilde a\|_1\,\radc\\
  &\le\bigl(\|a^\star-\hat a\|_1+\|\hat a-\tilde a\|_1\bigr)\radc\le2\epsilon\,\radc .
\end{aligned}
\end{equation}
The width form follows with $\epsilon=\Delta/2$. For the margin gap bound~\eqref{eq:thmC}, let
$j^\star$ be the dominant key and $\epsilon_m=2(n-1)e^{-\gamma}/(1+(n-1)e^{-\gamma})$ the radius
in the margin bound~\eqref{eq:margin}. By the attention intervals~\eqref{eq:aint},
$\alo_{j^\star}=(1+\sum_{l\ne j^\star}e^{\shi_l-\slo_{j^\star}})^{-1}$, and every exponent is at most
$-\gamma$ by the definition of the margin, so
\begin{equation*}
  \alo_{j^\star}\ge\bigl(1+(n-1)e^{-\gamma}\bigr)^{-1},
\end{equation*}
and hence every $a$ in the box and the simplex satisfies
\begin{equation*}
  \|a-e_{j^\star}\|_1=2(1-a_{j^\star})\le2(1-\alo_{j^\star})\le\epsilon_m .
\end{equation*}
The cut $\|a-e_{j^\star}\|_1\le\epsilon_m$ of the margin pair is therefore implied by the attention intervals, and the pattern polytope of the width pair is contained in that of the margin pair. The gap bound~\eqref{eq:thmB} applied to the margin pair gives the margin gap bound~\eqref{eq:thmC} for the certificate with the width pair.
\end{proof}

\begin{proof}[Proof of Proposition~\ref{prop:sharp}]
Take $d=3$, the fixed query $z_2=e_1$, which is also the second key, and the first key
$z_1=e_2+te_3$ with $|t|\le\rho$, so that $B_1=e_3$ and $\U_1=[-\rho,\rho]$. Let
\begin{equation*}
  M=\lambda\,e_1e_3^\top,
  \qquad
  N=-e_1(e_1+2e_3)^\top,
  \qquad
  c=e_1,
\end{equation*}
so that the value of a token $z$ along $c$ is $h(z)=-(e_1^\top z)-2(e_3^\top z)$. Then, $s_2=0$,
$s_1=\lambda t$, $h_2=-1$ and $h_1=-2t$, and the output is
\begin{equation}\label{eq:sharpphi}
  \varphi(t)=-1+\Lambda(\lambda t)\,(1-2t).
\end{equation}
Its derivative,
\begin{equation*}
  \varphi'(t)=\Lambda(\lambda t)\bigl[\lambda\bigl(1-\Lambda(\lambda t)\bigr)(1-2t)-2\bigr],
\end{equation*}
is negative for $\lambda\le\tfrac12$ and $|t|\le\tfrac12$, so
\begin{equation*}
  \pstar=\varphi(-\rho)=-1+\Lambda(-\lambda\rho)(1+2\rho).
\end{equation*}
The score intervals of the two keys are, respectively, $[-\lambda\rho,\lambda\rho]$ and $\{0\}$, so $\Delta=2\lambda\rho$, the
midpoint pattern is uniform, $\hmax=(2\rho,-1)$ and $\radc=\tfrac12(1+2\rho)$. The interval bound
$a_1\le\Lambda(\lambda\rho)$ is tighter than the width cut $a_1\le\tfrac12+\tfrac{\lambda\rho}2$, so
\begin{equation}\label{eq:sharprpp}
  \Rpp=-1+\Lambda(\lambda\rho)(1+2\rho),
  \qquad
  \Rpp-\pstar=(1+2\rho)\bigl(\Lambda(\lambda\rho)-\Lambda(-\lambda\rho)\bigr)=(1+2\rho)\tanh\tfrac{\lambda\rho}{2}.
\end{equation}
Since $\tfrac12\Delta\radc=\tfrac{\lambda\rho}{2}(1+2\rho)$ and $\tanh x=x(1+O(x^{2}))$, the gap
formula~\eqref{eq:sharp} follows.
\end{proof}

\begin{proof}[Proof of Proposition~\ref{prop:dual}]
For the width pair the first phase of the greedy transfer is void, because $\hat a$ lies in the box and on the simplex. The second phase moves mass from donors to receivers at a cost of two units of
budget per unit of mass. By the greedy solution of Proposition~\ref{prop:greedy}, the transfer is
optimal for all data. The order in which it visits the keys depends only on $\hmax$. After a small
change of $\epsilon$ or of $\ahi_j$ it therefore performs the same transfers, with the same saturated
keys and the same last receiver and donor, which lie strictly inside their intervals by assumption. The three
parts are proved in turn.
\begin{enumerate}[label=\normalfont(\roman*), leftmargin=*]
\item On the fixed polytope $\mathcal A$, $\Rpp$ is a maximum of functions linear in $\hmax$. Such a
maximum is convex, and it is differentiable where its maximizer is unique, with gradient equal to that
maximizer.
\item A change $\delta$ of $\epsilon$ changes the mass moved from $j^{-}$ to $j^{+}$ by $\delta/2$,
which is feasible for small $|\delta|$ of either sign because both keys lie strictly inside their
intervals, so
\begin{equation*}
  \Rpp(\epsilon+\delta)-\Rpp(\epsilon)=\tfrac12\bigl(\hmax_{j^{+}}-\hmax_{j^{-}}\bigr)\delta .
\end{equation*}
\item A change $\delta$ of $\ahi_j$ moves a mass $\delta$ between $j^{+}$ and $j$. The budget is
unchanged, because the deviations of $a_j$ and $a_{j^{+}}$ from $\hat a$ change by $\delta$ in
opposite directions, and the move is feasible for small $|\delta|$ because $j^{+}$ lies strictly inside
its interval. Receivers are filled in decreasing order of $\hmax$, so $\hmax_j\ge\hmax_{j^{+}}$ and
\begin{equation*}
  \Rpp\big|_{\ahi_j+\delta}-\Rpp\big|_{\ahi_j}=\bigl(\hmax_j-\hmax_{j^{+}}\bigr)\delta . \qedhere
\end{equation*}
\end{enumerate}
\end{proof}

\begin{proof}[Validity of $\Rlift$ and $\Rexp\le\Rpp$]
For an admissible input with base coordinates $u$ and pattern $a=a(z)$, let $X$ be the moment
matrix of $(u,a)$, and let $\tau=\lse(s(z))$ and $t_j=\log a_j$. Then, (R1) holds because
$\|u_k\|\le\rho_k$, and (R2) holds because $t_j=s_j-\tau=\log a_j$ lies in the log-attention
interval~\eqref{eq:tint}, $[\log\alo_j,\log\ahi_j]$, by the attention intervals of
Lemma~\ref{lem:intervals}(i). Constraint (R3) holds because $a_j=e^{t_j}\le\ell_j(t_j)$ and
$a\in[\alo,\ahi]$, and (R4) because $(a_j-\alo_j)(\ahi_j-a_j)\ge0$. Constraint (R5) holds because
\begin{equation*}
\begin{gathered}
  \|a_ju_j\|\le\rho_ja_j,
  \qquad
  \|(1-a_j)u_j\|\le\rho_j(1-a_j),
  \\
  \sum_ja_ju_k=u_k,
  \qquad
  \sum_ja_ja_l=a_l,
  \qquad
  a_ja_l\ge0,
\end{gathered}
\end{equation*}
(R6) because the pair is certified, and (R7) because the moment matrix of a single point is the rank-one matrix $(1,\xi)(1,\xi)^\top$. Moreover, $\Phi(X)=\varphi(z)$. For the comparison with
the first-order certificate, (R5) gives $g_j^\top Y[a_j,u_j]\le\|g_j\|\rho_j\bar a_j$, so that
\begin{equation*}
  \Phi(X)\le\sum_j\bar a_j\bigl(h_j(\bar z_j)+\rho_j\|g_j\|\bigr)=\sum_j\bar a_j\hmax_j,
\end{equation*}
while (R3) and (R6) place $\bar a$ in the pattern polytope.
\end{proof}

\begin{proof}[Proof of Theorem~\ref{thm:F0v}]
Let $X$, $\tau$ and $t$ be feasible for $\Rlift$, with first moments $\bar u,\bar a$, and let
$\Sigma= Y-\bar\xi\bar\xi^\top$ be the covariance matrix of $X$. It is the Schur complement of the entry $X_{00}=1$ in the moment matrix~\eqref{eq:moment}, whose first row is $(1,\bar\xi^\top)$ and whose lower right block is $Y$, and hence positive semidefinite by the moment conditions (R7). Write $h_j(u)$ for the value of key
$j$ along $c$ at the base coordinate $u$, so that $h_j(u)=h_j(\bar z_j)+g_j^\top u$. The proof has
four steps.

\emph{Step 1: the objective at the first moments.} The objective~\eqref{eq:liftedobj} is
\begin{equation*}
  \Phi(X)=\sum_j\Bigl(\bar a_j\,h_j(\bar z_j)+g_j^\top Y[a_j,u_j]\Bigr).
\end{equation*}
Splitting $Y[a_j,u_j]=\bar a_j\bar u_j+\Sigma[a_j,u_j]$ in it gives
\begin{equation}\label{eq:step1}
  \Phi(X)=\sum_j\bar a_j\,h_j(\bar u_j)+\sum_jg_j^\top\Sigma[a_j,u_j].
\end{equation}
By the Cauchy--Schwarz inequality for the positive semidefinite matrix $\Sigma$,
\begin{equation}\label{eq:step1cs}
  \bigl|g_j^\top\Sigma[a_j,u_j]\bigr|\le\sqrt{\Sigma[a_j,a_j]}\,\sqrt{g_j^\top\Sigma[u_j,u_j]g_j}.
\end{equation}
By (R4) and (R3), and by (R1),
\begin{equation*}
\begin{aligned}
  \Sigma[a_j,a_j]&\le(\bar a_j-\alo_j)(\ahi_j-\bar a_j)\le\tfrac14(\ahi_j-\alo_j)^{2},\\
  g_j^\top\Sigma[u_j,u_j]g_j&\le g_j^\top Y[u_j,u_j]g_j\le\|g_j\|^{2}\tr Y[u_j,u_j]\le\nu_{c,j}^{2}.
\end{aligned}
\end{equation*}
Hence,
\begin{equation}\label{eq:step1b}
  \Phi(X)\le\sum_j\bar a_j\,h_j(\bar u_j)+\frac12\sum_j\bigl(\ahi_j-\alo_j\bigr)\nu_{c,j}.
\end{equation}

\emph{Step 2: the softmax sandwich.} Let $\sigma=\sigma(t)$, which equals the softmax of the
lifted scores $\bar s(X)$ because $t=\bar s(X)-\tau$. By (R2), (R3) and the value-weighted sandwich identity of Proposition~\ref{prop:P7v} applied with $h_j=h_j(\bar u_j)$,
\begin{equation}\label{eq:step2}
  \sum_j\bar a_j\,h_j(\bar u_j)=\bar h_\sigma+\sum_jd_j\bigl(h_j(\bar u_j)-\bar h_\sigma\bigr),
\end{equation}
where $\bar h_\sigma=\sum_j\sigma_jh_j(\bar u_j)$, $0\le d_j\le D_j$ and $\sum_jd_j\le D_{\mathrm{tot}}$.

\emph{Step 3: the lifted scores.} By the expression~\eqref{eq:liftedscore}, the lifted score
differs from the score of the first moments by a covariance term, namely
\begin{equation*}
  \bar s_j(X)-s_j(\bar u)=\bigl\langle B_i^\top MB_j,\ \Sigma[u_i,u_j]\bigr\rangle .
\end{equation*}
For a positive semidefinite block matrix, the nuclear norm of an off-diagonal block is at most
the geometric mean of the traces of the diagonal blocks, so
\begin{equation}\label{eq:step3}
  \bigl|\bar s_j(X)-s_j(\bar u)\bigr|\le\|B_i^\top MB_j\|\sqrt{\tr\Sigma[u_i,u_i]\,\tr\Sigma[u_j,u_j]}\le\|B_i^\top MB_j\|\rho_i\rho_j\le\eta ,
\end{equation}
and the same bound holds for $j=i$. By the Lipschitz bound of Lemma~\ref{lem:lip},
$\|\sigma-\sigma(s(\bar u))\|_1\le\eta$. The first moments $\bar u$ form an admissible input by
(R1), with output $\varphi(\bar u)=\sum_j\sigma_j(s(\bar u))h_j(\bar u_j)$. Centring the values at
the midpoint $m$ of their range, so that $|h_j(\bar u_j)-m|\le\radc$,
\begin{equation}\label{eq:step3b}
  \bar h_\sigma-\varphi(\bar u)=\sum_j\bigl(\sigma_j-\sigma_j(s(\bar u))\bigr)\bigl(h_j(\bar u_j)-m\bigr)\le\eta\,\radc .
\end{equation}

\emph{Step 4: assembly.} From inequality~\eqref{eq:step3b},
\begin{equation*}
  \bar h_\sigma\le\pstar+\eta\,\radc,
  \qquad
  \bar h_\sigma\ge\varphi(\bar u)-\eta\,\radc\ge\phimin-\eta\,\radc .
\end{equation*}
Since $h_j(\bar u_j)\le\hmax_j$, every factor in the identity~\eqref{eq:step2} satisfies
$h_j(\bar u_j)-\bar h_\sigma\le\chi_j$, and since $d\ge0$ the sum in that identity is at
most $\sum_jd_j\chi_j\le\KN(\chi,D,D_{\mathrm{tot}})$. Combining with inequality~\eqref{eq:step1b},
\begin{equation}\label{eq:step4}
  \Phi(X)\le\pstar+\eta\,\radc+\KN(\chi,D,D_{\mathrm{tot}})+\frac12\sum_j\bigl(\ahi_j-\alo_j\bigr)\nu_{c,j}.
\end{equation}
Taking the supremum over feasible points gives the second-order bound~\eqref{eq:F0v}.

For boxes $\U_k=\{|u|\le \omega_k\}$, (R1) is replaced by the constraints
\begin{equation*}
  |\bar u_k|\le\omega_k,
  \qquad
   Y[u_k,u_k]_{ll}\le \omega_{k,l}^{2}.
\end{equation*}
The first moments are again admissible. The diagonal constraints give
$\tr\Sigma[u_k,u_k]\le\tr Y[u_k,u_k]\le\|\omega_k\|^{2}=\rho_k^{2}$, so Step~3 holds with the
circumradii, and since $| Y_{lm}|\le( Y_{ll} Y_{mm})^{1/2}$ for a positive semidefinite matrix,
\begin{equation}\label{eq:boxvar}
  g_j^\top Y[u_j,u_j]g_j\le\sum_{l,m}|g_{j,l}|\,|g_{j,m}|\,\omega_{j,l}\omega_{j,m}=\Bigl(\sum_l|g_{j,l}|\omega_{j,l}\Bigr)^{2}=\nu_{c,j}^{2},
\end{equation}
so Step~1 holds with the directional value width $\nu_{c,j}=\max_{|u|\le\omega_j}|g_j^\top u|$ of the
box.

Every step uses the moment matrix only through its first moments and its principal submatrices
indexed by the constant, the query, key $j$ and the weight of key $j$, together with (R1)--(R4).
The bound therefore also holds for the relaxation in which only these submatrices are required to
be positive semidefinite.
\end{proof}

\begin{proof}[Proof of Corollary~\ref{cor:orders}]
The second-order bound~\eqref{eq:F0v} of Theorem~\ref{thm:F0v} states that
\begin{equation*}
  \Rexp-\pstar\le\KN(\chi,D,D_{\mathrm{tot}})+\eta\,\radc+\frac12\sum_j\bigl(\ahi_j-\alo_j\bigr)\nu_{c,j}.
\end{equation*}
The two parts are proved in turn.
\begin{enumerate}[label=\normalfont(\roman*), leftmargin=*]
\item The output is a convex combination of values, so $\min_l\hmin_l$ is itself a certified lower
bound on $\min\varphi$, and the first-order certificate for $-c$ never returns a smaller one; one may
therefore assume $\phimin\ge\min_l\hmin_l$. Since $\hmax_j\le\max_l\hmax_l$, this gives $\chi_j\le(2+\eta)\radc$, and by the exact secant gap of Lemma~\ref{lem:secant}
\begin{equation*}
  \KN\le(2+\eta)\,\radc\min\Bigl\{\sum_jD_j,\ D_{\mathrm{tot}}\Bigr\}\le(2+\eta)\,\radc\min\{S,D_{\mathrm{tot}}\}.
\end{equation*}
The last term of the second-order bound is at most $\frac12\nu_cW$, with $W=\sum_j(\ahi_j-\alo_j)$, which gives the first inequality of the order bound~\eqref{eq:orderS}. For the second, $w_j\le2\Delta$ by the width bound~\eqref{eq:wbound}, so that, by the multiplicative interval bounds of Lemma~\ref{lem:intervals}(ii),
\begin{equation*}
  S\le\frac{(2\Delta)^{2}}{8}\sum_j\ahi_j\le\frac{\Delta^{2}}{2}e^{\Delta},
  \qquad
  W\le2\sinh\Delta .
\end{equation*}
\item The argument is in exact arithmetic. By the Cauchy--Schwarz inequality, each score satisfies
\begin{equation*}
  \bigl|s_j-\bar z_i^\top M\bar z_j\bigr|\le\|M\|\bigl(\rho\|\bar z_j\|+\rho\|\bar z_i\|+\rho^{2}\bigr),
\end{equation*}
so the score intervals, whether computed by this inequality or by the semidefinite bounds, which are
never wider, have width
\begin{equation*}
  \Delta\le2\|M\|(2R_0\rho+\rho^{2})\le1\qquad\text{for }\rho\le\rho_1 .
\end{equation*}
Moreover,
\begin{equation*}
  \eta\le\|M\|\rho^{2}\le\|M\|(2R_0\rho+\rho^{2})\le\tfrac12,
  \qquad
  \nu_c\le\beta_c\rho,
  \qquad
  \radc\le r_{0,c}+\nu_c\le r_{0,c}+\beta_c .
\end{equation*}
Substituting into the second inequality of the order bound~\eqref{eq:orderS}, namely
\begin{equation*}
  \Rexp-\pstar\le\radc\Bigl((2+\eta)\tfrac{\Delta^{2}}{2}e^{\Delta}+\eta\Bigr)+\nu_c\sinh\Delta ,
\end{equation*}
the inequalities
\begin{equation*}
  e^{\Delta}\le e,
  \qquad
  \Delta^{2}\le4\|M\|^{2}(2R_0+1)^{2}\rho^{2},
  \qquad
  \sinh\Delta\le\Delta\sinh1
\end{equation*}
gives
\begin{equation}\label{eq:radiusproof}
\begin{aligned}
  \Rexp-\pstar&\le(r_{0,c}+\beta_c)\Bigl(\tfrac52\cdot\tfrac e2\cdot4\|M\|^{2}(2R_0+1)^{2}+\|M\|\Bigr)\rho^{2}\\
  &\quad+2\sinh(1)\,\beta_c\|M\|(2R_0+1)\rho^{2}\le K_h\rho^{2}.
\end{aligned}
\end{equation}
For the first-order certificate, the gap bound of Theorem~\ref{thm:B} gives
\begin{equation*}
  \Rpp-\pstar\le\Delta\,\radc\le2\|M\|(2R_0+1)\rho\,(r_{0,c}+\beta_c)=K'_h\rho .
\end{equation*}
The bounds on the tightness radii follow, since a certificate whose gap is at most $K\rho^{q}$ on
$[0,\rho_1]$ is within $\varepsilon$ of the truth for all $\rho\le\min\{\rho_1,(\varepsilon/K)^{1/q}\}$.
On the two-key instances of Proposition~\ref{prop:sharp}, the gap of the first-order certificate is at least
$\frac{\lambda\rho}{4}$, because $\tanh x\ge x/2$ for $0\le x\le\frac12$, so it exceeds $\varepsilon$
as soon as $\rho>4\varepsilon/\lambda$. \qedhere
\end{enumerate}
\end{proof}

\begin{proof}[Proof of Proposition~\ref{prop:lower}]
Take $d=4$, the fixed query $z_2=e_1$, which is also the second key, and the first key
$z_1=e_2+u_1e_3+u_2e_4$ with $\|u\|\le\rho$, so that $B_1=[e_3\ e_4]$. Let
\begin{equation*}
  M=\lambda\,e_1e_3^\top,
  \qquad
  N=-e_1(e_1+2e_4)^\top,
  \qquad
  c=e_1 .
\end{equation*}
Then, $s_2=0$, $s_1=\lambda u_1$, $h_2=-1$ and $h_1=-2u_2$, and the output is
\begin{equation*}
  \varphi(u)=-1+\Lambda(\lambda u_1)(1-2u_2).
\end{equation*}
The score of the first key ranges over $[-\lambda\rho,\lambda\rho]$. Both attention weights
therefore have the interval $[\Lambda(-\lambda\rho),\Lambda(\lambda\rho)]$, and both log-attention
intervals have the width $w=\lambda\rho$, since $\log\Lambda(x)-\log\Lambda(-x)=x$. The width pair
is $\hat a=(\tfrac12,\tfrac12)$ and $\epsilon=\lambda\rho$, and $\eta=0$ because the query is fixed.

Let $u^\star$ be a worst-case input, $s^\star=\lambda u^\star_1$ and
$\sigma=(\Lambda(s^\star),\Lambda(-s^\star))$ its pattern. For $\delta\ge0$, let
$\bar a=(\sigma_1+\delta,\sigma_2-\delta)$, let $X$ be the moment matrix of the point
$(u^\star,\bar a)$, and let $\tau=-\log\bar a_2$ and $t=(s^\star-\tau,-\tau)$. Since $X$ is the
moment matrix of a single point with $\|u^\star\|\le\rho$ and $\bar a$ on the simplex, (R1), (R5)
and (R7) hold, and (R4) holds when $\bar a$ lies in its intervals. The lifted scores are the scores of
$u^\star$, and $\bar a_2=e^{t_2}$, while
\begin{equation*}
  e^{t_1}=e^{s^\star}(\sigma_2-\delta)=\sigma_1-e^{s^\star}\delta\le\bar a_1 .
\end{equation*}
The secant constraint of the second key holds because the secant lies above the exponential. That
of the first key, $\bar a_1\le\ell_1(t_1)$, is equivalent to
\begin{equation}\label{eq:lowerd}
  \delta\le\sigma_2\bigl(\ell_1(t_1)-e^{t_1}\bigr),
\end{equation}
because $1+e^{s^\star}=1/\sigma_2$. The remaining constraints, the intervals of $\bar a$, $t$ and
$\tau$ and the pattern cut (R6), hold for $\delta$ of order $\rho^{2}$ and small $\rho$, because
$|s^\star|$ stays below $\lambda\rho$ by a margin of order $\rho$, as shown next. The objective is
\begin{equation*}
  \Phi(X)=\bar a_1h_1(u^\star)+\bar a_2h_2=\pstar+\delta\bigl(1-2u^\star_2\bigr).
\end{equation*}

Since $\varphi(\rho v)=\varphi(0)+\rho\nabla\varphi(0)^\top v+O(\rho^{2})$ uniformly for
$\|v\|\le1$, every maximizer satisfies $\nabla\varphi(0)^\top u^\star/\rho\ge\|\nabla\varphi(0)\|-O(\rho)$.
As $\rho\to0$, therefore, $u^\star/\rho\to(\lambda,-4)/\sqrt{16+\lambda^{2}}$, the direction of the
gradient $(\lambda/4,-1)$ of $\varphi$ at the origin. Hence,
$s^\star/(\lambda\rho)\to\kappa=\lambda/\sqrt{16+\lambda^{2}}<1$, $\sigma\to(\tfrac12,\tfrac12)$
and $1-2u^\star_2\ge1$ for small $\rho$. At $\delta=0$ the log-attention $t_1=\log\Lambda(s^\star)$
lies at a relative position $x$, measured from the centre of its interval in units of the
half-width, with $x\to\kappa$, and for $0\le\delta\le\lambda^{2}\rho^{2}$ the relative position
moves by $O(\rho)$. Since the second derivative of the exponential is at least $e^{\tlo}$ on the
interval, the secant gap at relative position $x$ satisfies
\begin{equation*}
  \ell_1(t)-e^{t}\ge\tfrac12e^{\tlo}(t-\tlo)(\thi-t)=\tfrac{w^{2}}{8}e^{\tlo}\bigl(1-x^{2}\bigr).
\end{equation*}
Fix $\kappa'\in(\kappa,1)$ and choose
\begin{equation*}
  \delta=\sigma_2\,\tfrac{w^{2}}{8}\,e^{\tlo}\bigl(1-\kappa'^{2}\bigr),
\end{equation*}
which is of order $\rho^{2}$. For small $\rho$ it satisfies the condition~\eqref{eq:lowerd} and the
remaining constraints, so that
\begin{equation*}
  \liminf_{\rho\to0}\frac{\Rexp-\pstar}{\rho^{2}}\ge\lim_{\rho\to0}\frac{\delta}{\rho^{2}}=\frac{\lambda^{2}}{32}\bigl(1-\kappa'^{2}\bigr).
\end{equation*}
Letting $\kappa'\to\kappa$ and using $1-\kappa^{2}=16/(16+\lambda^{2})$ gives the lower
bound~\eqref{eq:lower}.

For the value-weighted secant slack, the first-order certificate for $-c$ gives
$\phimin=-\tfrac12+O(\rho)$ by the first-order gap bound of Theorem~\ref{thm:B}. Hence,
$\chi_1=\tfrac12+O(\rho)$ and $\chi_2=0$ for small $\rho$. By the expansion~\eqref{eq:Dasym} of the exact secant gap, $D=\frac{w^{2}}{8}e^{\tlo}(1+O(w))$, $D_1=\tfrac{1}{16}\lambda^{2}\rho^{2}\bigl(1+O(\rho)\bigr)$, while
$D_{\mathrm{tot}}=1-2\Lambda(-\lambda\rho)$ is of order $\rho$. The knapsack therefore puts the mass
$D_1$ on the first key, and $\KN(\chi,D,D_{\mathrm{tot}})=\chi_1D_1=\tfrac{1}{32}\lambda^{2}\rho^{2}\bigl(1+o(1)\bigr)$.
\end{proof}

\begin{proof}[Proof of Proposition~\ref{prop:projection}]
Let $\Pi=VV^\top$ be the orthogonal projection onto the row space of $P_h$, which is spanned by
the columns of $W_Q$, $W_K$ and $W_V$. Then,
\begin{equation*}
  M=\Pi M\Pi,
  \qquad
  N^\top c=\Pi N^\top c,
\end{equation*}
so every score, every value along $c$ and every lifted expression in
the lifted score~\eqref{eq:liftedscore} and the lifted objective~\eqref{eq:liftedobj} depends on a base coordinate $u$ only
through $V^\top u$, and on a second moment block $Y$ only through $V^\top Y V$. The map $V^\top$
sends the ball of radius $\rho_j$ in $\R^{d}$ onto the ball of radius $\rho_j$ in the reduced space,
because $\|V^\top u\|\le\|u\|$ and $V^\top(Vv)=v$ with $\|Vv\|=\|v\|$. The sets of reachable query, key and value vectors, and hence $\pstar$ and the support values $\hmax_j$, are therefore the same in both coordinate systems. For the relaxations, let
\begin{equation*}
  T=\diag\bigl(1,V^\top,\dots,V^\top,I_n\bigr).
\end{equation*}
Let $X$ be feasible in the original coordinates. Then, $TXT^\top$ is positive semidefinite, its first moments $V^\top\bar u_k$ have norm at most $\|\bar u_k\|$, and its diagonal blocks satisfy $\tr V^\top Y[u_k,u_k]V\le\tr Y[u_k,u_k]$ because $VV^\top\preceq I$. The norms in (R5) do not increase, and the linear constraints and the objective, which involve the base coordinates only through $V^\top$, are unchanged, so the reduced relaxation has at least the same value.
Conversely, if $X'$ is feasible in the reduced coordinates, $T^\top X'T$ is feasible in the original
ones with the same objective, because $V$ has orthonormal columns and preserves traces and
norms. The same two maps apply to the semidefinite score bounds.
\end{proof}

\begin{proof}[Proof of Proposition~\ref{prop:safe}]
For a feasible point and multipliers in the dual cones, every
$\lambda_k^\top\bigl(\mathcal A_k(X,x)+b_k\bigr)$ is nonnegative, so
\begin{equation*}
    \langle F,X\rangle+f^\top x+f_0\le f_{0,\lambda}+f_\lambda^\top x+\langle F_\lambda,X\rangle .
\end{equation*}
The linear term is at most $\sum_l\max\{(f_\lambda)_lx^{\mathrm{lo}}_l,(f_\lambda)_lx^{\mathrm{hi}}_l\}$. For the matrix
term, indices with $\delta_l=0$ have $X_{ll}=0$, hence a zero row and column, and the remaining
part of $X$ can be written as $\diag(\delta)^{1/2}\tilde X\diag(\delta)^{1/2}$ with $\tilde X\succeq0$ and $\tr\tilde X=\sum_lX_{ll}/\delta_l\le n_G$ by the group budgets. Then,
\begin{equation*}
    \langle F_\lambda,X\rangle=\bigl\langle\diag(\delta)^{1/2}F_\lambda\diag(\delta)^{1/2},\ \tilde X\bigr\rangle\le\lambda^{+}_{\max}\bigl(\diag(\delta)^{1/2}F_\lambda\diag(\delta)^{1/2}\bigr)\tr\tilde X .
  \qedhere
\end{equation*}
\end{proof}

\begin{proof}[Proof of Proposition~\ref{prop:T}]
The three parts are proved in turn.
\begin{enumerate}[label=\normalfont(\roman*), leftmargin=*]
\item Fix an operation $o$ and a sign vector $\varsigma$. The two points
$\bar x_o\pm\varsigma\circ\omega_o$ are opposite corners of $\mathcal B_o$, where $\ell_o\ge\lambda_or_o$; since
$\ell_o$ is affine,
\begin{equation*}
  \ell_o(\bar x_o)=\tfrac12\bigl(\ell_o(\bar x_o+\varsigma\circ\omega_o)+\ell_o(\bar x_o-\varsigma\circ\omega_o)\bigr),
\end{equation*}
which is at least the bracket in the lower bound~\eqref{eq:Tlower}. Also,
$\ell_o(\bar x_o)\ge\lambda_or_o(\bar x_o)=0$. Since $0\in\mathcal D$ and $x_o(0)=\bar x_o$ for every
$o$, $U$ is at least the value of its objective at $\delta=0$, which is $\sum_o\ell_o(\bar x_o)$.
For a unary operation, Taylor's theorem gives
\begin{equation*}
  \lambda_or_o(\bar x_o\pm\omega_o)=\tfrac12\lambda_o\phi_o''(\xi_\pm)\,\omega_o^{2}\ge\tfrac12\underline\kappa_o\,\omega_o^{2};
\end{equation*}
for a product, $r_o(\bar x+a,\bar y+b)=ab$, so the bracket equals
$\lambda_o\varsigma_1\varsigma_2\omega_{o,x}\omega_{o,y}$, whose maximum over the signs is
$|\lambda_o|\omega_{o,x}\omega_{o,y}$.
\item For $\delta\in\mathcal D$ the segment from $\bar x_o$ to $x_o(\delta)$ lies in $\mathcal B_o$, so
Taylor's theorem with Lagrange remainder along it gives
\begin{equation*}
  \lambda_or_o\bigl(x_o(\delta)\bigr)=\tfrac12(G_o\delta)^\top\lambda_o\nabla^{2}\phi_o(\xi)(G_o\delta)\le\tfrac12\delta^\top G_o^\top K_oG_o\delta .
\end{equation*}
Summing over $o$ gives $\Psi(\delta)\le\frac12\delta^\top\mathcal Q\delta$, and the maximum of a quadratic
form over a ball of radius $\rho$ is $\rho^{2}\lambda_{\max}(\mathcal Q)_+$. If every $\phi_o$ is
quadratic and $K_o=\lambda_o\nabla^{2}\phi_o$, Taylor's formula is exact and
$\Psi=\frac12\delta^\top\mathcal Q\delta$.
\item Back-substitution starts from the output $\Psi(\delta)$, written as a linear function of the
outputs of the last operations, and replaces each output $\phi_o(x_o)$, with its current
multiplier $\hat\lambda_o$, by an affine function $\hat\lambda_o\phi_o(x_o)\le a_o(x_o)$ valid on $I_o$. It ends with an affine function $A\delta+b$ of the inputs. Each replacement changes the
expression, evaluated at the true values of all intermediate quantities, by exactly
$a_o(x_o(\delta))-\hat\lambda_o\phi_o(x_o(\delta))$, so that
\begin{equation*}
  A\delta+b=\Psi(\delta)+\sum_o\bigl[a_o\bigl(x_o(\delta)\bigr)-\hat\lambda_o\phi_o\bigl(x_o(\delta)\bigr)\bigr]
  \qquad\text{for every }\delta\in\mathcal D .
\end{equation*}
At $\delta=0$, where $\Psi(0)=0$ and every operand is nominal, the right side is
$\sum_o\ell_o(\bar x_o)$ with
\begin{equation*}
  \ell_o(x)=a_o(x)-\hat\lambda_o\bigl(\phi_o(\bar x_o)+\nabla\phi_o(\bar x_o)^\top(x-\bar x_o)\bigr),
\end{equation*}
an affine function with $\ell_o\ge\hat\lambda_or_o$ on $I_o$. The argument of part~(i) applied to
each $\ell_o$ at the corners $\bar x_o\pm\varsigma\circ\omega_o\in I_o$, together with
$U=\max_{\mathcal D}(A\delta+b)\ge b$, gives the claim.
\end{enumerate}
\end{proof}

\begin{proof}[Proof of Corollary~\ref{cor:T}]
With $\phi_o$ quadratic and $K_o=\lambda_o\phi_o''=\underline\kappa_o$, the quadratic upper bound of Proposition~\ref{prop:T}(ii) is an equality, $\Psi(\delta)=\frac12\delta^\top\mathcal Q\delta$, whose maximum
over the ball is $\frac12\rho^{2}\lambda_{\max}(\mathcal Q)$ since $\mathcal Q\succeq0$. Part~(i) with
$\omega_o=\rho\|g_o\|$ gives
\begin{equation*}
  U\ge\sum_o\tfrac12\underline\kappa_o\rho^{2}\|g_o\|^{2}=\tfrac12\rho^{2}\tr\mathcal Q .
\end{equation*}
The Shor relaxation maximizes $\frac12\langle\mathcal Q,X\rangle$ over $X\succeq0$ with
$\tr X\le\rho^{2}$; its value is $\frac12\rho^{2}\lambda_{\max}(\mathcal Q)$, attained at
$X=\rho^{2}vv^\top$ for a unit top eigenvector $v$. Finally, for $\mathcal Q\succeq0$,
\begin{equation*}
  \lambda_{\max}(\mathcal Q)\le\tr\mathcal Q\le\rank(\mathcal Q)\,\lambda_{\max}(\mathcal Q). \qedhere
\end{equation*}
\end{proof}

\begin{proof}[Proof of Lemma~\ref{lem:chain}]
Formula~\eqref{eq:H0} is the second-order chain rule for a computational graph in reverse form~\cite{griewank2008evaluating}. Affine maps have no second derivative, and a node
$v=\phi_o(x_o)$ contributes its adjoint times its local Hessian contracted with the first
derivatives of its operands, which is $\lambda_o\kappa_og_og_o^\top$ for a unary node and
$\lambda_o(g_{o,x}g_{o,y}^\top+g_{o,y}g_{o,x}^\top)$ for a product. For any vectors $a,b$ and $t>0$,
\begin{equation*}
  \pm\bigl(ab^\top+ba^\top\bigr)\preceq t\,aa^\top+t^{-1}bb^\top,
\end{equation*}
since the difference is
\begin{equation*}
  \bigl(\sqrt t\,a\mp b/\sqrt t\bigr)\bigl(\sqrt t\,a\mp b/\sqrt t\bigr)^\top\succeq0 .
\end{equation*}
Choosing
$t=\|g_{o,y}\|/\|g_{o,x}\|$ for each product, and dropping the unary terms with
$\lambda_o\kappa_o<0$, gives $H_0\preceq\bar H$ for a positive semidefinite $\bar H$ with
$\frac12\rho^{2}\tr\bar H=s_{\mathrm{op}}$. Then,
\begin{equation*}
  \lambda_{\max}(H_0)_+\le\tr(H_0^{+})=\max_{0\preceq\Pi\preceq I}\tr(\Pi H_0)\le\max_{0\preceq\Pi\preceq I}\tr(\Pi\bar H)=\tr\bar H .
\end{equation*}
\end{proof}

\section{Relaxation Details}\label{app:relaxation}
\setcounter{equation}{0}

This appendix writes out the level-one relaxation of Definition~\ref{def:R1} in full, for ball, ellipsoid and box token sets, and records the implementation choices on which the experiments rest.

Let $K_p\subseteq J$ be the set of perturbed tokens, each with base coordinates
$u_k\in\R^{r_k}$, and let the fixed tokens have $r_k=0$. With biases absorbed into the
centres, the score of key $j$ and the value of key $j$ along $c$ are, respectively,
\begin{equation}\label{eq:app-affine}
  s_j(u)=s^{0}_j+(p^{q}_j)^\top u_i+(p^{k}_j)^\top u_j+u_i^\top Q_ju_j,
  \qquad
  h_j(u_j)=h^{0}_j+g_j^\top u_j,
\end{equation}
with $Q_j=B_i^\top MB_j$, $p^{q}_j=B_i^\top M\bar z_j$, $p^{k}_j=B_j^\top M^\top\bar z_i$,
$s^{0}_j=\bar z_i^\top M\bar z_j$ and $h^{0}_j=c^\top N\bar z_j$. For $j=i$ the two linear terms act on the same coordinate. The
lifted vector is $\xi=\bigl((u_k)_{k\in K_p},a_1,\dots,a_n\bigr)$, of dimension
$n_\xi=\sum_kr_k+n$, and $X$ is its moment matrix~\eqref{eq:moment}, of side $n_\xi+1$, whose
first row is $(1,\bar\xi^\top)$ and whose lower right block $Y$ holds the second moments. The
scalars $\tau$, $t_1,\dots,t_n$ and the variables $p,q\in\R^{n}$ of the pattern cut are
not lifted.

For balls and ellipsoids in base coordinates, $\U_k=\{\|u\|\le\rho_k\}$, the relaxation
maximizes
\begin{equation}\label{eq:app-obj}
  \Phi(X)=\sum_{j\in J}\Bigl(h^{0}_j\,\bar a_j+g_j^\top Y[a_j,u_j]\Bigr)
\end{equation}
subject to the following constraints, in which the lifted score is
\begin{equation*}
  \bar s_j(X)=s^{0}_j+(p^{q}_j)^\top\bar u_i+(p^{k}_j)^\top\bar u_j+\bigl\langle Q_j, Y[u_i,u_j]\bigr\rangle .
\end{equation*}
\begin{enumerate}[(R1)]
\item $\|\bar u_k\|\le\rho_k$ and $\tr Y[u_k,u_k]\le\rho_k^{2}$ for $k\in K_p$.
\item $\slo_j\le\bar s_j(X)\le\shi_j$, $t_j=\bar s_j(X)-\tau$, $\tlo_j\le t_j\le\thi_j$ for
$j\in J$, and $\taulo\le\tau\le\tauhi$.
\item $(\bar a_j,1,t_j)\in\Kexp$, $\bar a_j\le e^{\tlo_j}+m_j(t_j-\tlo_j)$ with
$m_j=(e^{\thi_j}-e^{\tlo_j})/w_j$ and $w_j=\thi_j-\tlo_j$ (omitted when $w_j=0$), $\alo_j\le\bar a_j\le\ahi_j$, and
$\sum_j\bar a_j=1$.
\item $Y[a_j,a_j]\le(\alo_j+\ahi_j)\bar a_j-\alo_j\ahi_j$.
\item $\| Y[a_j,u_j]\|\le\rho_j\bar a_j$ and $\|\bar u_j- Y[a_j,u_j]\|\le\rho_j(1-\bar a_j)$
for $j\in K_p$; $\sum_j Y[a_j,u_k]=\bar u_k$ for $k\in K_p$; $\sum_j Y[a_j,a_l]=\bar a_l$ and
$Y[a_j,a_l]\ge0$ for all $j,l$.
\item $\bar a-\hat a=p-q$, $p,q\ge0$ and $\sum_j(p_j+q_j)\le\epsilon$. The certified pair is
the width pair, with $\hat a=\sigma(s^{\mathrm{mid}})$ and $\epsilon=\Delta/2$, or, in the
experiments, the margin pair when it gives the smaller radius. The margin pair is the one-hot
pattern of the dominant key with the radius $2(n-1)e^{-\gamma}/(1+(n-1)e^{-\gamma})$ of the margin
bound~\eqref{eq:margin}. The cut of the margin pair is implied by the attention intervals in (R3), as the proof of the first-order gap bound, Theorem~\ref{thm:B}, shows. On those instances the experiments therefore drop the cut in effect, and keeping the width cut instead would never give a weaker relaxation.
\item $X\in\Kpsd^{n_\xi+1}$ and $X_{00}=1$.
\end{enumerate}
The norm constraints are second-order cones, and the exponential cone constraint is
$\bar a_j\ge e^{t_j}$. For the query, which is also key $i$, the product $Y[a_i,u_i]$ is the
product of its own weight with its own perturbation, as for every other key.

For a box $\U_k=\{|u|\le \omega_k\}$, with $\omega_k\in\R^{r_k}_{+}$ and inequalities between
vectors understood componentwise, (R1) and the products with the sets in (R5) become
\begin{equation}\label{eq:app-box}
  |\bar u_k|\le\omega_k,\quad
   Y[u_k,u_k]_{ll}\le \omega_{k,l}^{2},\quad
  | Y[a_j,u_j]|\le\omega_j\,\bar a_j,\quad
  |\bar u_j- Y[a_j,u_j]|\le\omega_j(1-\bar a_j),
\end{equation}
the second family being the products $(\omega_{k,l}-u_{k,l})(\omega_{k,l}+u_{k,l})\ge0$. In an
ablation on synthetic heads, the products of the box constraints of pairs of coordinates, the reformulation--linearization constraints on $Y[u_j,u_k]$, were also added. They did not change the order of accuracy. The proof of the second-order bound, Theorem~\ref{thm:F0v}, for boxes uses only
the constraints~\eqref{eq:app-box} and the box variance bound~\eqref{eq:boxvar}, namely
$g_j^\top Y[u_j,u_j]g_j\le\nu_{c,j}^{2}$.

Keys with negligible attention are pruned. For keys with $\ahi_j\le10^{-6}$ the
exponential cone and secant constraints are removed, while their score, log-attention and
attention intervals, the definition $t_j=\bar s_j(X)-\tau$, the interval diagonal and the products
with the sets are kept. This removes badly scaled cones, with $t_j$ of the order
of $-18$, that caused the solver to stop short at high temperature. Removing constraints
enlarges the feasible set, so the certificate remains valid. The proof of the second-order bound, Theorem~\ref{thm:F0v}, goes through with $d_j=\bar a_j-e^{t_j}$ of either sign for a pruned key. The unpruned keys still satisfy
$\sum_jd_j\le D_{\mathrm{tot}}=1-\sum_j\alo_j$, because $\bar a_j\ge\alo_j$ for the pruned ones. Since $\bar a_j$
and $e^{t_j}\le e^{\thi_j}$ both lie in $[0,\ahi_j]$, a pruned key has $|d_j|\le\ahi_j$. Every value
$h_l(\bar u_l)$ lies in the interval $[\min_l\hmin_l,\max_l\hmax_l]$, of length $2\radc$, which
also contains $\bar h_\sigma=\sum_l\sigma_lh_l(\bar u_l)$ with $\sigma=\sigma(t)$. The pruned keys
therefore add at most $2\radc\sum_{j\ \text{pruned}}\ahi_j\le2n\cdot10^{-6}\radc$ to the right side
of the second-order bound~\eqref{eq:F0v}, namely to
\begin{equation*}
  \KN(\chi,D,D_{\mathrm{tot}})+\eta\,\radc+\frac12\sum_j\bigl(\ahi_j-\alo_j\bigr)\nu_{c,j}.
\end{equation*}

The safe dual bound of Proposition~\ref{prop:safe} is applied to the relaxation as it is passed to the solver. The multipliers are the solver's dual values projected onto the dual cones. Multipliers of linear inequalities are clipped at zero, and the scalar part of a second-order cone multiplier is raised to the norm of its vector part. An exponential cone multiplier is moved to the boundary of the dual cone in the component that certifies membership. Since any multipliers in the dual cones give a valid bound, both sign conventions of the returned duals are evaluated and the smaller bound is kept. The a priori bounds of Proposition~\ref{prop:safe} are those implied by the relaxation itself. They are the ball constraints for the first moments of the base coordinates, the intervals for $\bar a$, $\tau$ and $t$, and $p,q\in[0,\epsilon]$. For the diagonal of $X$, each ball block forms a group with $\delta_l=\rho_k^{2}$ and budget one, each box coordinate has
$\delta_l=\omega_{k,l}^{2}$, each weight has $\delta_l=(\ahi_j)^{2}$ by (R4), and $X_{00}=1$. The matrix
term is bounded by the smaller of the eigenvalue term
$n_G\,\lambda^{+}_{\max}(\diag(\delta)^{1/2}F_\lambda\diag(\delta)^{1/2})$ of the safe
bound~\eqref{eq:safe} and the entrywise bound
$\sum_{l,m}|(F_\lambda)_{lm}|(\delta_l\delta_m)^{1/2}$ of the positive semidefinite part. Rounding errors in
the evaluation of the bound~\eqref{eq:safe}, including the eigenvalue computation, are covered by
adding $64\,\mathbf u$ times the sum of the absolute values of all its terms, where
$\mathbf u=2^{-53}$ is the unit roundoff. This allowance is generous for the sizes involved, but it is not a proof in the sense of interval arithmetic. A fully rigorous evaluation would follow VSDP~\cite{jansson2007rigorous}. The same remark applies to the score intervals. Their endpoints are computed by the Shor relaxation at the solver's default tolerance of $10^{-9}$ and widened by
$10^{-7}\max\{1,|v|\}$ for an endpoint of value $v$.

The score intervals themselves are computed as follows. For a key $j\ne i$, the upper end $\shi_j$ is the value of the Shor relaxation of $\max\{s_j(u):\|u_i\|\le\rho_i,\ \|u_j\|\le\rho_j\}$, that is, the maximum of the lifted score over moment matrices of $(u_i,u_j)$ subject to (R1). The lower end is obtained from $-s_j$. Because the score depends on the query only through
$W_Q^\top z_i$ and on the key only through $W_K^\top z_j$, the relaxation is solved in coordinates
of dimension $2d_k$, a matrix of side $129$ on GPT-2, with the same value. For $j=i$ the problem has one ball, and the relaxation is exact by the S-lemma. The attention intervals then follow in closed form from Lemma~\ref{lem:intervals}, namely
\begin{equation*}
  \alo_j=\Bigl(1+\sum_{l\ne j}e^{\shi_l-\slo_j}\Bigr)^{-1},
  \qquad
  \ahi_j=\Bigl(1+\sum_{l\ne j}e^{\slo_l-\shi_j}\Bigr)^{-1},
\end{equation*}
and the definition~\eqref{eq:tint} gives the interval $[\lse(\slo),\lse(\shi)]$ of the normalizer
and the log-attention intervals $[\log\alo_j,\log\ahi_j]$. The log-sum-exp is evaluated in a numerically stable way. The exact secant gaps follow from the closed form~\eqref{eq:Dclosed}, $D_j=e^{\tlo_j}((1+y)\log(1+y)-y)$ with $y=(e^{w_j}-1)/w_j-1$. For narrow intervals, this formula subtracts nearly equal numbers. It is therefore computed with the standard library functions $\operatorname{expm1}(x)=e^{x}-1$ and $\operatorname{log1p}(x)=\log(1+x)$, which remain accurate for small $x$, and with a short Taylor series when $w$ is very small.

A node of the branch-and-bound is the instance with the score intervals of some keys
restricted. Since a child is a restriction of its parent, its bound is the minimum of its parent's bound and the safe value of the relaxation with the node's score intervals and all controls recomputed from them. Nodes are explored best first. A child is discarded when the solver
certifies that its relaxation is infeasible or when its bound does not exceed the best feasible
value. The search stops when the bracket is below $10^{-3}$ of the output's range of motion or after $24$ semidefinite solves. The certificate is then the minimum of the first-order certificate and the largest bound of the open leaves. Feasible values, which certify the lower ends of all brackets, are obtained by local ascent from many starts. The starts are the nominal tokens, the maximizers of the support values used by the first-order
certificate, and the first moments of the relaxation with roundings of its three leading
eigenvectors. They also include the first moments of every branch-and-bound leaf and $24$ random
points on the boundaries of the token sets. Each
ascent runs a limited-memory quasi-Newton method in base coordinates, with projection onto the
token sets, for at most $2{,}000$ iterations.

The relaxations of the synthetic experiments are solved through a modelling layer with the
solver's default tolerances of $10^{-9}$. On GPT-2 they are passed directly to the solver's task
interface, with primal and dual feasibility tolerances of $10^{-6}$ and a relative gap tolerance of $10^{-8}$. A unit is re-solved at $10^{-9}$ whenever the difference between its safe value and the
solver's dual objective exceeds $1\%$ of the gap between the certificate and the best
feasible value, and the smaller safe value is kept. The exact references on synthetic heads are nonconvex programs in the log form of the softmax. They have the bilinear scores and the products of weights and values as explicit constraints, the exponential as a nonlinear constraint handled by spatial branch-and-bound, and all controls of Table~\ref{tab:controls} as variable bounds. They are
solved to a relative gap of $10^{-6}$ or to the time limit, $300$ seconds for $\rho\le0.2$ and
$90$ seconds for larger radii, in which case only their
bounds are used.

The GPT-2 experiments use the following five prompts, numbered from zero as in
Section~\ref{sec:experiments}.
\begin{enumerate}[label=\arabic*., start=0, leftmargin=*]
\item The small dog ran across the green field to find the red ball that the children had thrown.
\item Every morning the baker opens the shop at six, and the smell of fresh bread fills the quiet
street.
\item When the train arrived late, the passengers waited on the cold platform and talked about the
weather.
\item Anna wrote a long letter to her brother, and then she walked to the post office to send it.
\item The river flows past the village, under the stone bridge, and into the lake where the boats
are kept.
\end{enumerate}

\end{appendices}

\end{document}